%% file: main.tex
\documentclass[11pt]{article}

\usepackage[utf8]{inputenc}
\usepackage{amsmath, amssymb}
\usepackage{graphicx}
\usepackage{caption}
\usepackage{subcaption}
\usepackage{algorithm}
\usepackage{algpseudocode}
\usepackage{multirow}
\usepackage{array}
\usepackage{listings}
\usepackage{tabularx}
\usepackage{xcolor}
\usepackage{geometry}
\usepackage{booktabs}
\usepackage{hyperref}
\newcommand\ainote[1]{\textcolor{red}{[AI: #1]}}

\title{\textbf{Towards High-Fidelity Synthetic Multi-platform Social Media Datasets via Large Language Models}}

\author{
Henry Tari$^1$, Nojus Šereiva$^1$, Rishabh Kaushal$^{1,2}$, Thales Bertaglia$^3$, Adriana Iamnitchi$^1$ \\
$^1$Maastricht University, Netherlands \\
$^2$Indira Gandhi Delhi Technical University for Women, India \\
$^3$Utrecht University, Netherlands
}

\date{2 May 2025}

\begin{document}
\maketitle
\begin{abstract}
Social media datasets are essential for research on a variety of topics, such as disinformation, influence operations, hate speech detection, or influencer marketing practices. However, access to social media datasets is often constrained due to costs and platform restrictions. Acquiring datasets that span multiple platforms, which is crucial for understanding the digital ecosystem, is particularly challenging.
This paper explores the potential of large language models to create lexically and semantically relevant social media datasets across multiple platforms, aiming to match the quality of real data. We propose multi-platform topic-based prompting and employ various language models to generate synthetic data from two real datasets, each consisting of posts from three different social media platforms. We assess the lexical and semantic properties of the synthetic data and compare them with those of the real data.
Our empirical findings show that using large language models to generate synthetic multi-platform social media data is promising, different language models perform differently in terms of fidelity, and a post-processing approach might be needed for generating high-fidelity synthetic datasets for research. In addition to the empirical evaluation of three state of the art large language models, our contributions include new fidelity metrics specific to multi-platform social media datasets. 
\end{abstract}

\vspace{1em}

\input{sections/introduction}
\input{sections/related-work}
\input{sections/datasets}
\input{sections/new-methodology}
\input{sections/new-evaluations}
\input{sections/summary}
\bibliographystyle{plain}
\bibliography{references}

\end{document}

%% file: sections/introduction.tex
\section{Introduction}

Social media platforms have enabled the development of many socio-economic processes over the last decade. From connecting people worldwide to promoting goods and services through influencer marketing, these processes were analyzed in quantitative and qualitative studies by a multi-disciplinary community that includes sociologists, political scientists, communication experts, economists, and computer scientists. Such studies enabled progress in understanding individual and at-scale behaviors, detecting malicious actors and operations, identifying political manipulation and disinformation campaigns, modeling information diffusion processes, etc. At the core of all these studies is data from social media platforms, which typically record how users interact with content and each other. Never perfect~\cite{Tufekci_2014} and often costly to collect and curate~\cite{doi:10.1126/science.aaz8170}, social media datasets, in general, cannot be shared with the research community due to privacy and legal concerns even when social media posts are public. 

With the emergence of different social platforms with different types of content and audiences, it became clear that no one platform accurately reflects society on a particular topic at any time. Consequently, research shifted towards multi-platform studies of election discussions~\cite{10.1145/3328529.3328562,10.1145/3583780.3615121}, coordinated information operations~\cite{doi:10.1080/10584609.2019.1661889}, or cyberbullying~\cite{10.1007/s10579-020-09488-3}. However, datasets from multiple social media platforms are even more challenging to collect and curate and much more difficult to re-collect for reproducibility studies or ``hydrate'' from message identifiers, as the individual platform challenges are compounded. For example, publicly available Twitter datasets that include only the tweet IDs (and no actual content) are much more expensive to hydrate now due to policy changes at X related to its research API. Moreover, in most platforms, users remove their posts or accounts, or accounts are blocked: even when data collection is still practical, it is not guaranteed that the same data can be retrieved at a different time, even when using the same collection parameters. 

This paper poses the following question: \emph{Given a dataset comprising of text messages from various social media platforms, how accurately can we generate a corresponding synthetic dataset?} 
Specifically, we focus on multi-platform social media dataset that include messages posted publicly by a select group of users or pertaining to certain topics over a specified period. 
 Such messages are typically collected due to shared topics, such as elections~\cite{Aiyappa2023} or geo-political events~\cite{DBLP:conf/icwsm/NgHI21_}, or are collected based on authorship (i.e., all messages posted by a specified group of users during a time interval on all considered platforms). In addition to capturing the diversity of posting practices (for example, heavy use of emojis typical to Instagram and TikTok), and the audiences and functionalities specific to each platform, it is the connection between the simultaneous conversations on the different platforms via shared lexical entities (keywords, URLs, hashtags, user tags), shared users, and shared narratives that are often important features for problems such as detecting multi-platform coordinated information operations~\cite{wilson21cross-platformiops,DBLP:conf/misdoom/BarberoCKGSI23}.  
 
Our long-term goal is to facilitate reproducibility in social media research, as outlined in~\cite{schoch2024computational}, without contravening legal or platform-specific regulations. Ideally, researchers should be able to create and publicly share an accurate synthetic dataset derived from a multi-platform social media dataset gathered for their studies. This dataset could then be used with similar reliability as real data for various purposes, including training machine learning algorithms, testing tools on diverse datasets, and conducting reproducibility assessments. We hypothesize that this method will, in the long run, enhance transparency regarding content management on social media platforms, an initiative now supported by the EU Digital Services Act, thereby offering better protection against systemic risks. 

We take the first steps towards this vision by making the following contributions that significantly extend our previous work. 
First, we improve our methodology in generating synthetic multi-platform datasets by introducing a specialized prompting approach (that we call Multi-Platform Topic Model). We compare this new approach with our previously evaluated prompting technique proposed in ~\cite{tari2024leveraging} and discover its advantages and limitations. 
Second, we extend our empirical evaluation to a second dataset collected though a different process (that is, authorship-based vs. topic-based). This second dataset also expands our coverage of social media platforms to six very popular platforms: Twitter, Instagram, Facebook, TikTok, YouTube and Reddit.
Third, we compare three state-of-the-art commercial Large Language Models and show the differences in the data they generate. 
And finally, we introduce new metrics of data fidelity that capture the multi-platform essence of the datasets we want to generate. 

%% file: sections/related-work.tex
\section{Related Work}
\label{sec:related_work}
Recent developments in artificial intelligence, particularly in Large Language Models (LLMs), have been adopted in social science research. LLMs have been used to simulate human-like behaviors~\cite{grossmann2023ai}, assist in survey research~\cite{jansen2023employing}, help in human subject studies~\cite{aher2023using}, act as proxies for human samples~\cite{argyle2023out}, generate synthetic human-computer interaction data~\cite{hamalainen2023evaluating} and create artificial social media data~\cite{tornberg2023simulating}. However, we focus on the role of LLMs in generating realistic social media data in multi-platform scenarios. 
In the first sub-section, we discuss works that leverage LLMs to augment social media datasets and study the efficacy on downstream task.
Given that we propose a new prompt technique, in the second sub-section, we study prior works to understand different types of prompting techniques. 

\subsection{Role of LLMs in social media data augmentation and downstream task performance}
Data-driven machine learning requires large labeled training data.
However, obtaining high-quality labeled data is often a challenge.
Traditionally, numerous techniques \cite{li2022data} have been proposed to augment textual data, for instance, back-translation \cite{sennrich2015improving} and EDA \cite{wei2019eda}.
Researchers have studied the role of LLMs to generate synthetic dataset. 
Moller et al.~\cite{moller2023prompt} found that synthetic GPT-4 and Llama-2 data improves rare class performance in multi-class classification tasks. They evaluated their approach to ten tasks in computational social science. To generate synthetic data, they prompted LLMs by giving examples from single social media platform datasets.
Li et al.~\cite{li2023synthetic} studied the effectiveness of synthetic data produced by LLMs on ten different classification tasks. They found a negative association between the subjectivity of classification as measured at the instance level and task level and model performance. Models trained on synthetic data underperformed for all the classification tasks. 
Models performed comparably for less subjective tasks and poorly for highly subjective tasks.
They observed that a few short examples of prompting are better than zero prompting. 
More specifically,
LLMs are being used to generate social media data to improve the performance of a specific downstream task.
Bertaglia et al.~\cite{bertaglia2024instasynth} generated synthetic Instagram posts using four prompting strategies to augment training data for the sponsored content detection task. They observed that the fidelity of the generated synthetic dataset and its utility for the downstream task can conflict. 
Ghanadian et al.~\cite{ghanadian2024socially} investigated the feasibility of generating synthetic Reddit posts for suicidal ideation. LLMs were prompted to generate suicidal text related to different factors like depression, anxiety, anger, hopelessness, and more. They used two prompt approaches: zero-shot (with no examples) and few-shot (with two examples for each factor).
They found that combining synthetic with real datasets gives the best results. 
Hartvigsen et al.~\cite{hartvigsen2022toxigen} created ToxiGen, a dataset of toxic and benign statements synthetically generated using GPT-3 by giving examples from Reddit in their prompt. They observed improvement in toxicity detection using ToxiGen data. 
In contrast, Schmidhuber et al.~\cite{schmidhuber2024llm} performed experiments on six toxicity-related datasets and found that hate classification is not improved across all.
Similarly, Das et al.~\cite{das2024offlandat} proposed a dataset comprising implicit offensive speech targeted against 38 groups. They used zero-shot prompting grounded with positive intention to enable chatGPT to produce offensive text. Both ~\cite{hartvigsen2022toxigen} and ~\cite{das2024offlandat} employed human evaluation to assess the quality of the generated text.
Veselovsky et al.~\cite{veselovsky2023generating} focused on the sarcasm detection task using self-disclosed sarcastic tweets. They proposed three prompting approaches to generate faithful synthetic sentences, namely, grounding (by giving examples), taxonomy-based (theorizing a concept), and filtering (fine-tuning a model to distinguish synthetic from real sentences and then using it on LLM outputs to filter out sentences labeled as synthetic). They found that grounding is most effective in detecting sarcasm. 
Dos et al.~\cite{dos2024identifying} used LLMs to augment existing tweet data for identifying issues raised by citizens to improve policymaking. 
Bucur et al.~\cite{bucur2024leveraging} augment social media posts on Reddit using GPT-3.5 and LLama-3 to improve richness in data in terms of diverse ways to express emotions for depression detection. 
Wagner et al.~\cite{wagner2024power} demonstrated improved stance detection using LLM-generated data from Mistral-7B. They produce synthetic data for specific debate questions and find that fine-tuning the models with synthetic data produces better results. They further improved performance by fine-tuning models using the most information samples for which stance detection models are most uncertain. 
Kazemi et al.~\cite{kazemi2025synthetic} leveraged three LLMs, GPT-4o, Llama-3.3, and Grok-2, to label and generate synthetic data for cyberbullying detection. BERT-based models trained on synthetic data performed comparable to models trained on real data. 
Pastor et al.~\cite{pastor2025enhancing} exhibited comparable performance for both top-down and bottom-up parsers in discourse parsing using LLM-generated synthetic social media data on Twitter and Reddit.
Ignat et al.~\cite{ignat2024cross} used few-shot prompting to generate 2k inspiring synthetic Reddit posts using GPT-4 for inspiration detection in social media posts. They compared inspirational posts across two different cultures, India and the UK. They found a noticeable difference in how posts are composed using topic modeling. Further, machine learning models can easily detect posts across cultures.
Dunngood et al.~\cite{dunngood} studied the presence of emoji in GPT-3.5 generated social media content. They found that human-generated content mostly has emojis of hearts, faces, and hands, whereas LLM-generated content has a broader distribution of emojis. 
Most of these studies focus on using LLMs to augment data with an objective of improving performance of downstream task. In contrast, we focus on fidelity of synthetic datasets generated by our proposed prompting approach.  

\subsection{Prompt Approaches for Data Augmentation}
With the emergence of LLMs, numerous works~\cite{liu2023pre} have emerged that leverage different prompt approaches, most popular being zero-shot or few-shot prompting.
Ubani et al.~\cite{ubani2023zeroshotdataaug} used chatGPT with zero-shot prompting to create synthetic data for low-resource settings. They evaluated three datasets related to movie reviews, spoken natural language for personal services, and question classification. 
Dai et al.~\cite{dai2023auggpt} proposed an approach for text augmentation by using single-turn and multi-turn dialogues in chatGPT. They rephrased the input sentence into multiple sentences similar in concept but different in semantics. They validated their approach on three datasets - Amazon customer reviews, common medical symptom descriptions, and scientific abstracts.
Aboutalebi et al.~\cite{aboutalebi2024magid} introduced an approach to generate synthetic multi-modal datasets. They provide a text-only dialogue (sequence of utterances) to LLMs and prompt LLMs to identify the most suitable textual utterance that can be augmented with an image to make the dialogue more engaging. They used zero-shot, few-shot, and chain of reasoning techniques to identify the most suitable utterances. 
Josifoski et al.~\cite{josifoski2023exploiting} used LLMs with zero-shot and few-shot prompting to generate input sentences for a known target triplet comprising of subject, object, and their relation from a well-known Wikidata knowledge graph dataset.
Most of these prior works use a random example for few-shot prompting. In contrast, we propose a topic model based selection of examples to be fed as input to the LLMs. 

%% file: sections/datasets.tex
\section{Datasets}

For this study, we selected two multi-platform datasets, each containing messages posted by users on three different platforms. 
The first is a subset of the dataset collected by Aiyappa et al.~\cite{Aiyappa2023} on the US elections 2022 from Twitter, Facebook, and Reddit.  
We used a sample of the original multi-platform dataset related to the 2022 US midterm elections 
containing posts from Twitter, Facebook and Reddit published between October 1 and December 25, 2022. The 2022 midterm US elections were held on November 8, 2022.
We refer to this dataset as \emph{the election dataset}. 
The second consists of posts on TikTok, Instagram, and YouTube by social media influencers collected by Gui et al.~\cite{gui2024across}. 
These influencers are registered with the Dutch Media Authority based on registration obligations, as of July 1st, 2022~\cite{MinOCW2022}. We refer to this dataset as \emph{the influencers dataset}.

These two datasets cover a set of characteristics that enable a broader exploration of synthetic social media datasets. 
First, both are multi-platform datasets, connected by either users authoring posts on multiple platforms (the Dutch influencers) or by topic (the US 2022 elections). 
Second, taken together, they cover six popular social media platforms. 
Third, the topics covered and the discussion styles are diverse. 
On the one hand, the US election dataset should be more homogeneous in topic and yet more heterogeneous in authorship, since it contains posts from users with various lexical styles and political leanings. 
On the other hand, the Dutch influencer dataset is more diverse in topics (e.g., due to the different brands and images promoted) and potentially more homogeneous in authorship style since this user group has the common goal of monetizing their content. 

\begin{table}[htbp]
\centering
\caption{Platform Traits on Datasets. Average number of hashtags, mentions, URLs, and emojis per post on different platforms in elections and influencers dataset samples.}
\resizebox{\textwidth}{!}{%
\begin{tabular}{lcccc|lcccc}
\toprule
\multicolumn{5}{c}{\textbf{Elections Dataset}} & \multicolumn{5}{|c}{\textbf{Influencers Dataset}}\\
\hline
  \textbf{Platform} & \textbf{Hashtags} & \textbf{Mentions} & \textbf{URLs} & \textbf{Emojis}  & \textbf{Platform} & \textbf{Hashtags} & \textbf{Mentions} & \textbf{URLs} & \textbf{Emojis} \\
\midrule 
 Facebook  &  0.98 & 0.05 & 0.21 & 0.00   & Instagram  & 2.65 & 0.56 & 0.03 & 1.95 \\
  Twitter  &  1.37 & 0.73 & 0.40 & 0.40 & TikTok & 2.88 & 0.29 & 0.00 & 1.28\\
  Reddit  & 0.06 & 0.01 & 0.28 & 0.01 & YouTube & 0.91 & 0.51 & 11.08 & 2.64\\
\bottomrule
\end{tabular}
}
\label{tab:basic_platform_traits}
\end{table}
  
In this work, we focus on textual data only, and thus ignore user identities, user interactions, and the timing of the posts. 
Given their different focus on content modalities, posts from different platforms are highly distinct in length and usage of textual features such as hashtags and URLs. Table~\ref{tab:basic_platform_traits} shows that hashtags are hardly used in Reddit but common in Twitter, Instagram, and TikTok; Mentions are highly popular on Twitter followed by Instagram and YouTube and practically non-existent on Reddit; URLs are heavily present on YouTube video descriptions and non-existent on TikTok and Instagram. Finally, emojis are mostly prevalent in YouTube, followed by Instagram and TikTok and only on Twitter in the elections dataset.

%% file: sections/new-methodology.tex
\section{Multi Platform Synthetic Data Generation via LLMs}
\label{sec:topic-prompt}

\begin{figure}[!h]
\centering
{\includegraphics[width=0.8\linewidth]{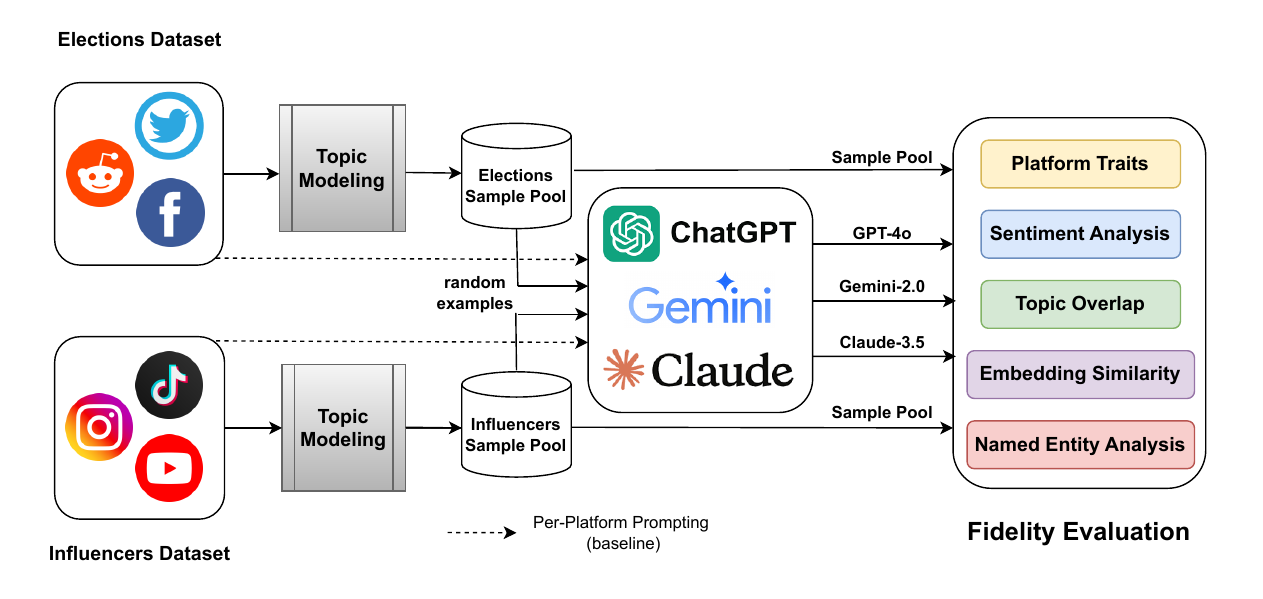}}
\caption{Multi Platform Topic Modeling (MPTM)-based prompting approach for generating and evaluating synthetic social media datasets. Posts from different platforms from two datasets are first processed using topic models to identify shared topics. Based on these topics, we create sample pools for each platform. From these pools, we randomly select examples to use for LLM prompting. For comparison, our baseline method (\textit{Per-Platform prompting}, presented in our previous work \cite{tari2024leveraging}) selects random examples directly from each platform. This baseline approach is illustrated with dotted lines.}


\label{fig:prompt_pipeline}
\end{figure}

Figure~\ref{fig:prompt_pipeline} illustrates our strategy for generating synthetic data. We first select example posts from each platform in the real dataset based on shared topics, forming the sample pool shown in the figure. A random subset of these sample pools is then used as input examples for the LLMs. We evaluate three widely used LLMs available at the time of this study: OpenAI's GPT-4o, Google's Gemini 2.0 Flash, and Anthropic's Claude 3.5 Haiku.
These models present either the most cost-effective current generation options 
or, in the case of GPT-4o, offer vastly improved performance~\cite{paech2023eq} over their counterparts. 
Even though the chosen Anthropic and Google models are not their best offerings, they still score well on relevant benchmarks and offer a better balance between price and performance than the provider's benchmark models, thus being preferable for large dataset generation.

We evaluate the fidelity of the generated synthetic datasets by comparing them to the original real datasets using four metrics. First, we analyze the frequency of lexical entities that are specific to each platform, such as hashtags, emojis, and URLs. For example, hashtags and emojis are commonly used on Instagram but are rare on Reddit, while URLs are frequent on YouTube but almost never appear on TikTok. This metric helps us assess whether platform-specific language patterns are preserved in the synthetic data.

Second, we compare the sentiment distributions in the real and synthetic datasets. In our previous work~\cite{tari2024leveraging}, we observed that GPT 3.5-turbo tended to generate overly positive content. Here, we extend the analysis to different LLMs to check whether they capture the sentiment characteristics of each platform; for instance, the relatively negative tone often found on Reddit versus the generally more positive tone of Instagram

Third, we assess topic overlap to evaluate the effectiveness of our topic-based prompting strategy for multi-platform datasets. This metric checks whether topics present on multiple platforms in real data are represented similarly in synthetic data. Additionally, we measure the semantic similarity between real and synthetic data using embedding similarity metrics. 

Finally, we perform a named entity analysis. We build bipartite graphs where one set of nodes represents posts and the other represents named entities, with edges indicating the mention of an entity in a post. We use this structure to analyze the presence of common entities across real and synthetic datasets, showing how well named entity distributions are preserved. 

\subsection{Multi-Platform Topic Model based Prompting (MPTM)}

In our previous work~\cite{tari2024leveraging}, we generated synthetic social posts for each platform separately by providing the LLM with platform-specific examples. This approach resulted in a disconnected collection that did not represent the overlap in topics that is specific to multi-platform datasets. 
In particular, topics that appeared across multiple platforms were significantly underrepresented in the synthetic datasets compared to the real data~\cite{tari2024leveraging-arxiv}. 
In this work, we address this limitation by using a prompting method that incorporates topics shared across multiple social media platforms, allowing for the generation of more coherent and interconnected synthetic datasets. 

We describe our proposed Multi-Platform Topic Model (MPTM) prompting approach below and summarize it in Algorithm~\ref{algo:topic_prompt}.

\begin{enumerate}
    \item Posts from multiple platforms are considered documents and fed into the BERTopic model. 
    \item BERT model (all-MiniLM-L6-v2
    ) is used to generate post embedding vectors.
    \item UMAP (Uniform Manifold Approximation and Projection) is applied to reduce the dimensionality of these embedding vectors, creating a more compact representation while preserving the semantic relationships.
    \item HDBSCAN clustering algorithm is then used to group similar vectors into clusters based on the reduced embeddings.
    \item C-TFIDF is computed for each cluster to extract topic-representative keywords within that cluster.
    \item From these topics ($T$), a random topic of interest is selected.
    \item For the selected topic, a sample pool ($SP$) of social media posts is created by selecting $m$ posts from each platform related to the selected topic.
    \item Once Sample Pool ($SP$) is created, we perform few-shot example prompting in which example posts are selected from $SP$.
\end{enumerate}

This approach ensures that we obtain a diverse yet topically coherent sample of posts across multiple platforms. It is important to note that the posts in the Sample Pool ($SP$) reflect popular topics drawn from multiple platforms, without being associated with the specific platform on which they were originally published. Consequently, the outputs generated by the LLMs through MPTM prompting do not correspond to any single platform, but instead represent a blend of content from different platforms. 

\begin{algorithm}
\caption{Multi-Platform Topic Modeling (MPTM) based Prompting}
\label{algo:topic_prompt}
\begin{algorithmic}[1]
\Require Posts across $n$ multiple platforms $D_{1} = \{p^1_1, p^1_2, \ldots, p^1_{m_1}\}$; $D_{2} = \{p^2_1, p^2_2, \ldots, p^2_{m_2}\}$ ... $D_{n} = \{p^n_1, p^n_2, \ldots, p^n_{m_n}\}$
\Ensure Topics $T = \{t_1, t_2, \ldots, t_k\}$ and Sample Pool $SP = p_1,p_2,.....$

\Function{GetTopics}{$D_1, D_2, .... D_n$}

\State // Step 0: Combine posts from all platforms
$D = D_1 + D_2 + ... + D_n$
\State // Step 1: Generate Embeddings
\ForAll{$p_i \in D$}
    \State $e_i \gets \text{BERTEmbed}(p_i)$
\EndFor
\State $E \gets \{e_1, e_2, \ldots, e_n\}$

\State // Step 2: Dimensionality Reduction
\State $E_{low} \gets \text{UMAP}(E)$

\State // Step 3: Clustering
\State $C \gets \text{HDBSCAN}(E_{low})$

\State // Step 4: Topic Keyword Extraction
\ForAll{cluster $c_j \in C$}
    \State $D_j \gets \text{documents in } c_j$
    \State $TFIDF_j \gets \text{c-TFIDF}(D_j)$
    \State $t_j \gets \text{Top keywords from } TFIDF_j$
\EndFor
\State \Return $T = \{t_1, t_2, \ldots, t_k\}$
\EndFunction

\Function{GetSample}{$T = \{t_1, t_2, \ldots, t_k\}$, $D = \{D_1, D_2, ..., D_n\}, C, m$}
\State $t_r \gets \text{RandomSelect}(T)$
\State $SP \gets \emptyset$ 
\ForAll{$D_i \in D$}
    \State $P_i^r \gets \{p \in D_i \mid \text{ClusterOf}(p) = \text{ClusterOf}(t_r)\}$
    \State $S_i \gets \text{RandomSample}(\min(|P_i^r|, m), P_i^r)$ 
    \State $SP \gets SP \cup S_i$
\EndFor

\State \Return $SP$ =  \{$p_1,p_2, \ldots, p_m\}$
\EndFunction

\end{algorithmic}
\end{algorithm}

\subsection{Prompt Design}
We use random example posts from the topic-coherent sample pool of social media posts for elections and influencer datasets. The following is the prompt template that we use in all MPTM experiments:

\begin{verbatim}
You are a social media content generator. I'm going to show you some example posts, 
and I want you to generate new posts that are similar in style, tone, and content,
but not identical. The generated posts should look like they could come from
the same types of users and platforms as the examples.
Here are some examples of the kind of posts I'm looking for:
{
Example 1;
Example 2;
Etc...
}
Please generate 5 new posts that are similar to these examples. Make them realistic
and varied.The posts should be about similar topics and use similar language 
patterns as well as be of similar length. Return just the posts without any 
additional commentary.
\end{verbatim}
We use the following prompt template for all per-platform prompting experiments:

\begin{verbatim}
Using following examples, generate 2 new social media posts, keeping them true to 
their content and writing style.

        Example 1: {0}
        Example 2: {1}
        Example 3: {2}
        
Consider the three examples I gave you and generate the two outputs according to 
them 
\end{verbatim}

The LLM output is then parsed and cleaned. This ensures that we do not save erroneous outputs and recover from incorrect return format outputs.
We chose to increase the prompt size due to the extended context windows of new models, which gives the LLMs more information on which to base their responses. The 9-5 ratio was chosen as it provided the maximum amount of context for the model, with a decent input-to-output conversion while not exceeding the token limits set by the models or their APIs.

%% file: sections/new-evaluations.tex
\section{The Text Fidelity of Synthetic Social Media Posts}
\label{sec:results}
We evaluate the text fidelity of the synthetic social media posts across four dimensions. First, we examine the presence of platform-specific textual features commonly found in real social media posts, including hashtags, user tags, URLs, and emojis. These elements help capture the stylistic and functional differences between platforms. Second, we compare the sentiment of synthetic posts to that of real posts to assess whether language models reflect platform-specific tone and emotional expression. Third, we evaluate topic diversity and coherence by analyzing how well the synthetic posts represent the range and structure of topics found in the original datasets. Finally, we measure the semantic similarity between real and synthetic posts using embeddings to capture deeper lexical and contextual alignment.

\begin{table}[htbp]
\centering
\caption{Platform Traits of Real and Synthetic Social Media Data for Elections and Influencers Dataset. Original Data: Social media posts that are fed to the topic model as input; MPTM prompt: Synthetic social media posts generated from our proposed Multi-Platform Topic Modeling based prompting; Baseline results of per-platform prompting (platform aware prompting from our previous work~\cite{tari2024leveraging}).}
\resizebox{\textwidth}{!}{%
\begin{tabular}{llrrrr|llrrrr}
\toprule
\multicolumn{6}{c}{\textbf{Elections Dataset}} & \multicolumn{6}{|c}{\textbf{Influencers Dataset}}\\
\hline
\textbf{Folder} & \textbf{Provider} & \textbf{Hashtags} & \textbf{Mentions} & \textbf{URLs} & \textbf{Emojis} & \textbf{Folder} & \textbf{Provider} & \textbf{Hashtags} & \textbf{Mentions} & \textbf{URLs} & \textbf{Emojis} \\
\midrule 
\multirow{3}{*}{Original Data} & Facebook  &  0.98 & 0.05 & 0.21 & 0.00  & \multirow{3}{*}{Original Data} & Instagram  & 2.65 & 0.56 & 0.03 & 1.95 \\
  & Twitter  &  1.37 & 0.73 & 0.40 & 0.40 & & TikTok & 2.88 & 0.29 & 0.00 & 1.28\\
  & Reddit  & 0.06 & 0.01 & 0.28 & 0.01 & & YouTube & 0.91 & 0.51 & 11.08 & 2.64\\
  \midrule
  
  \midrule
\multirow{3}{*}{MPTM prompt} & Sample Pool & 0.76 & 0.26 & 0.28 & 0.13 & \multirow{3}{*}{MPTM prompt} & Sample Pool & 2.14 & 0.50 & 3.57 & 2.03 \\
  & Claude-3.5 & 1.14 & 0.30 & 0.05 & 0.23 & & Claude-3.5 & 2.45 & 0.80 & 0.63 & 2.83 \\
  &  Gemini-2.0 & 1.77 & 0.18 & 0.04 & 0.05 & & Gemini-2.0 & 4.19 & 0.46 & 0.54 & 1.76 \\
  &  GPT-4o & 0.66 & 0.29 & 0.20 & 0.96 & & GPT-4o & 1.62 & 0.58 & 0.38 & 6.85  \\
\midrule

\midrule
\multirow{4}{*}{Facebook} & Claude-3.5 & 1.18 & 0.03 & 0.04 & 0.20 & \multirow{4}{*}{Instagram}  & Claude-3.5 &  0.80 & 0.38 & 0.01 & 1.94 \\
  & Gemini-2.0 & 0.99 & 0.02 & 0.05 & 0.00 & & Gemini-2.0 & 1.91 & 0.42 & 0.02 & 1.31 \\
  & GPT-4o & 0.52 & 0.05 & 0.13 & 0.49 & & GPT-4o & 1.09 & 0.34 & 0.02 & 3.10\\
  & Facebook  & 0.90 & 0.05 & 0.20 & 0.01 & & Instagram  & 2.62 & 0.54 & 0.03 & 1.94  \\
\midrule
\multirow{4}{*}{Twitter} & Claude-3.5  & 1.32 & 0.57 & 0.09 & 0.37 & \multirow{4}{*}{TikTok} & Claude-3.5  & 1.73 & 0.32 & 0.00 & 1.46\\
  & Gemini-2.0 & 1.52 & 0.63 & 0.14 & 0.15 & & Gemini-2.0 & 2.66 & 0.23 & 0.00 & 0.99\\
  & GPT-4o &  0.98 & 0.61 & 0.39 & 1.25 &   & GPT-4o   & 1.60 & 0.26 & 0.00 & 2.71 \\
  & Twitter &  1.41 & 0.72 & 0.41 & 0.39 & & TikTok  & 2.84 & 0.30 & 0.00 & 1.29 \\
\midrule
\multirow{4}{*}{Reddit} & Claude-3.5 & 0.95 & 0.00 & 0.00 & 0.32 & \multirow{4}{*}{YouTube} & Claude-3.5 & 1.22 & 0.39 & 2.21 & 2.44 \\
  & Gemini-2.0  & 0.51 & 0.00 & 0.02 & 0.01  & & Gemini-2.0 &  1.24 & 0.24 & 3.02 & 1.27 \\
  & GPT-4o &  0.09 & 0.01 & 0.06 & 0.39 & & GPT-4o & 0.58 & 0.51 & 1.94 & 6.20\\
  & Reddit  & 0.04 & 0.01 & 0.27 & 0.02 & & YouTube  & 0.93 & 0.51 & 10.82 & 2.57 \\
\bottomrule
\end{tabular}
}
\label{tab:platform_traits}
\end{table}

\subsection{Hashtags, User Tags, URLs and Emojis in Synthetic Data}
Table~\ref{tab:platform_traits} presents the average number of hashtags, user tags, URLs, and emojis in real and synthetic posts for both Multi-Platform Topic Modeling (MPTM) based prompting and per-platform prompting.
Organizing social media posts around hashtags is common on platforms like Twitter, Instagram, and TikTok, as observed from the original data. On average, GPT-4o tends to create fewer hashtags in synthetic data generated by MPTM prompting and for per-platform prompting for all platforms except Reddit, where hashtags in reality are much less used. 
This result contrasts with our previous study~\cite{tari2024leveraging}, where GPT-3.5-Turbo generated a higher number of hashtags in synthetic data, particularly for Facebook and Twitter.
However, hashtags, except for Instagram and TikTok, are generally overrepresented in the synthetic data generated by Claude-3.5 and Gemini-2.0.  
On social media platforms, hashtags are meant to connect posts with similar topics, yet understandably, GPT-4o underplays this characteristic feature. 
This difference was also observed by Bertaglia et al.~\cite{bertaglia2024instasynth} and likely relates to the sequential manner in which LLMs work without trying to make connections between individual outputs. 
However, this suggests that prompting strategies emphasizing shared hashtags among the examples given in prompts could lead to more hashtag-rich generated datasets.

Mentioning other users on social media platforms is common on Twitter, Instagram, and YouTube, but is almost entirely absent on Reddit and only very sparsely used on Facebook and TikTok. 
In MPTM prompting, we observe that Claude-3.5 overestimates the frequency of user mentions, while Gemini-2.0 underestimates it, across both elections and influencers datasets. In contrast, GPT-4o matches the mention patterns found in real data.
In the per-platform prompting strategy for Twitter, all models generate fewer mentions than those present in the real tweets, consistent with our previous findings~\cite{tari2024leveraging}. Similar trends are seen for Instagram and YouTube, except for GPT-4o, which matches the frequency of mentions in YouTube video descriptions. This result shows an improvement over GPT-3.5-Turbo, which previously failed to match mention patterns of real data~\cite{tari2024leveraging}.  


YouTube video descriptions contain a large number of URLs, as influencers often link to other platforms. Twitter, Reddit, and Facebook follow in URL frequency.
In multi-platform settings, no LLM matches the volume of URLs found in real social media data, especially due to YouTube's dominance. This underrepresentation persists in per-platform prompting, largely because other platforms (in both elections and influencer datasets) include significantly fewer URLs. As a result, the gap between real and synthetic data is limited. These URL-related patterns align with previous findings~\cite{tari2024leveraging}.

Emojis in social media posts often indicate emotional expressions. Remarkably, there is a clear difference in emoji usage between the two datasets: the influencers dataset contains many emojis, while posts related to the US elections contain significantly fewer.  Across platforms, emojis are most frequently used on YouTube, followed by Instagram, TikTok, and Twitter.
It is unclear whether this difference is primarily driven by the nature of the topics or by platform-specific norms. However, in the synthetic data generated using MPTM prompting, GPT-4o significantly overestimates emoji usage across both datasets. Claude-3.5 also tends to produce more emojis than observed in the real data. In contrast, Gemini-2.0 consistently underestimates emoji frequency.
In per-platform prompting, GPT-4o produces more emojis across all platforms for both datasets, showing a shift in behavior from GPT-3.5-Turbo~\cite{tari2024leveraging}, where higher emoji usage was limited to Instagram, TikTok, and Twitter. Similar to the multi-platform settings, Gemini-2.0 continues to generate fewer emojis in all per-platform prompts, while Claude-3.5 either matches or exceeds the emoji counts compared to real data.

\begin{table}[htbp]
\centering
\caption{Sentiment analysis of real and synthetic social media data. Original Data: Social media posts that are fed to the topic model as input; MPTM: Multi-Platform Topic Modeling based prompting to generate synthetic social media posts; Baseline results of per-platform prompting (platform aware from Tari et al.~\cite{tari2024leveraging}). Negative sentiment content in downplayed in synthetic datasets.}
\resizebox{\textwidth}{!}{%
\begin{tabular}{llrrr|llrrr}
\toprule
\multicolumn{5}{c}{\textbf{Elections Dataset}} & \multicolumn{5}{|c}{\textbf{Influencers Dataset}}\\
\hline
\textbf{Folder} & \textbf{Provider} & \textbf{Positive} & \textbf{Neutral} & \textbf{Negative}  & \textbf{Folder} & \textbf{Provider} & \textbf{Positive} & \textbf{Neutral} & \textbf{Negative} \\
\midrule
\multirow{3}{*}{Original Data} & Facebook  & 23.4 & 45.0 & 31.6 &  \multirow{3}{*}{Original Data} 
  & Instagram  & 55.2 & 34.4 & 10.4 \\
  & Twitter  & 28.8 & 32.8 & 38.4 & & TikTok  & 38.3 & 46.9 & 14.8\\
  & Reddit  & 5.8 & 21.4 & 72.8 & & YouTube  & 12.7 & 84.8 & 2.5 \\
  \midrule
  
  \midrule
\multirow{3}{*}{MPTM prompt} & Sample Pool & 17.2 & 34.2 & 48.6 & \multirow{3}{*}{MPTM prompt} & Sample Pool & 34.8 & 56.4 & 8.9 \\
  & Claude-3.5 & 27.4 & 17.4 & 55.2 & & Claude-3.5 & 69.7 & 20.3 & 10 \\
  & Gemini-2.0 & 28.4 & 16.4 & 55.2 & & Gemini-2.0 & 68.8 & 22.6 & 8.6 \\
  & GPT-4o & 37.4 & 20.2 & 42.4 & & GPT-4o & 80.7 & 15.6 & 3.7\\
\midrule

\midrule
\multirow{4}{*}{Facebook} & Claude-3.5 & 50.8 & 28.9 & 20.3 & \multirow{4}{*}{Instagram} & Claude-3.5 & 78.8 & 16.9 & 4.3 \\
  & Gemini-2.0 & 39.4 & 35.6 & 25.0 & & Gemini-2.0 & 68.3 & 25.7 & 6.0 \\
  & GPT-4o & 50.6 & 32.8 & 16.6 & & GPT-4o & 74.9 & 23.2 & 1.9\\
  & Facebook  & 23.9 & 44.5 & 31.6 & & Instagram  & 55.9 & 35.4 & 8.8  \\
\midrule
\multirow{4}{*}{Twitter} & Claude-3.5 & 44.6 & 22.0 & 33.4 & \multirow{4}{*}{TikTok} & Claude-3.5 & 53.4 & 33.4 & 13.1 \\
  & Gemini-2.0 & 35.8 & 28.2 & 36.0  & & Gemini-2.0 & 48.8 & 36.7 & 14.5\\
  & GPT-4o & 49.9 & 26.8 & 23.3  & & GPT-4o & 50.0 & 45.0 & 5.0\\
  & Twitter & 26.3 & 35.4 & 38.3 & & Tiktok  & 38.7 & 47.8 & 13.5\\
\midrule
\multirow{4}{*}{Reddit} & Claude-3.5 & 10.2 & 16.3 & 73.4 & \multirow{4}{*}{YouTube} & Claude-3.5 & 49.5 & 46.9 & 3.6\\
  & Gemini-2.0 & 12.3 & 18.5 & 69.2 &  & Gemini-2.0 & 36.4 & 57.9 & 5.7\\
  & GPT-4o & 18.7 & 15.7 & 65.6  & & GPT-4o & 37.6 & 61.3 & 1.1 \\
  & Reddit  & 5.9 & 20.8 & 73.3 & & YouTube  & 13.6 & 83.2 & 3.2 \\
\bottomrule
\end{tabular}
}
\label{tab:sentiments}
\end{table}

\subsection{Sentiment Analysis} 

Moving beyond lexical tokens, we compare sentiments expressed in generated content with real data in Table~\ref{tab:sentiments}. 
Sentiment analysis is a relevant task for evaluating the fidelity of synthetic social media data because it captures the emotional tone that characterizes communication across different platforms. We use sentiment distributions as a fidelity metric, specifically by measuring and comparing the sentiments expressed in synthetic datasets against those in real datasets in multi-platform and per-platform prompting.
We employed the 
\emph{cardiffnlp/twitter-XLM-roBERTa-base-sentiment}
proposed by Loureiro et al.~\cite{loureiro-etal-2022-timelms} available at Hugging Face\footnote{https://huggingface.co/cardiffnlp/twitter-roberta-base-sentiment-latest} to classify posts as \textit{positive}, \textit{negative}, or \textit{neutral}.
The model is based on the RoBERTa architecture~\cite{Liu2019RoBERTaAR} and is pre-trained for language modeling on ~198M tweets and then fine-tuned for sentiment analysis on the TweetEval benchmark~\cite{barbieri2020tweeteval}.
The model provides a weighted score distribution for each post across positive, neutral, and negative sentiment. We used the sentiment with the highest score in the final calculations.
Table~\ref{tab:sentiments} presents the percentage of posts belonging to a particular sentiment category for each experiment.
We observe that posts in the elections dataset, particularly those on Reddit, have a high proportion of negative sentiments, followed by Twitter and Facebook. 
In contrast, the influencers dataset posts are mostly positive or neutral across all platforms. 
For MPTM prompting, we observe that GPT-4o (in line with GPT-3.5-Turbo~\cite{tari2024leveraging}) underestimates the negative sentiments 
whereas both Claude-3.5 and Gemini-2.0 overestimate the negative sentiment in generated elections posts.
In the influencer MPTM results, Gemini-2.0 slightly underestimates the negative sentiment, while Claude-3.5 again shows a negative bias.
In per-platform prompting, GPT-4o continues to  underestimate negativity for all platforms, with Gemini-2.0 and Claude-3.5 following a similar trend with the exception of Reddit and YouTube for Claude-3.5 and YouTube with TikTok for Gemini-2.0 model.

\subsection{Topic Generation and Overlap}
\begin{figure}[htb]
    \centering
    \includegraphics[width=0.75\linewidth]{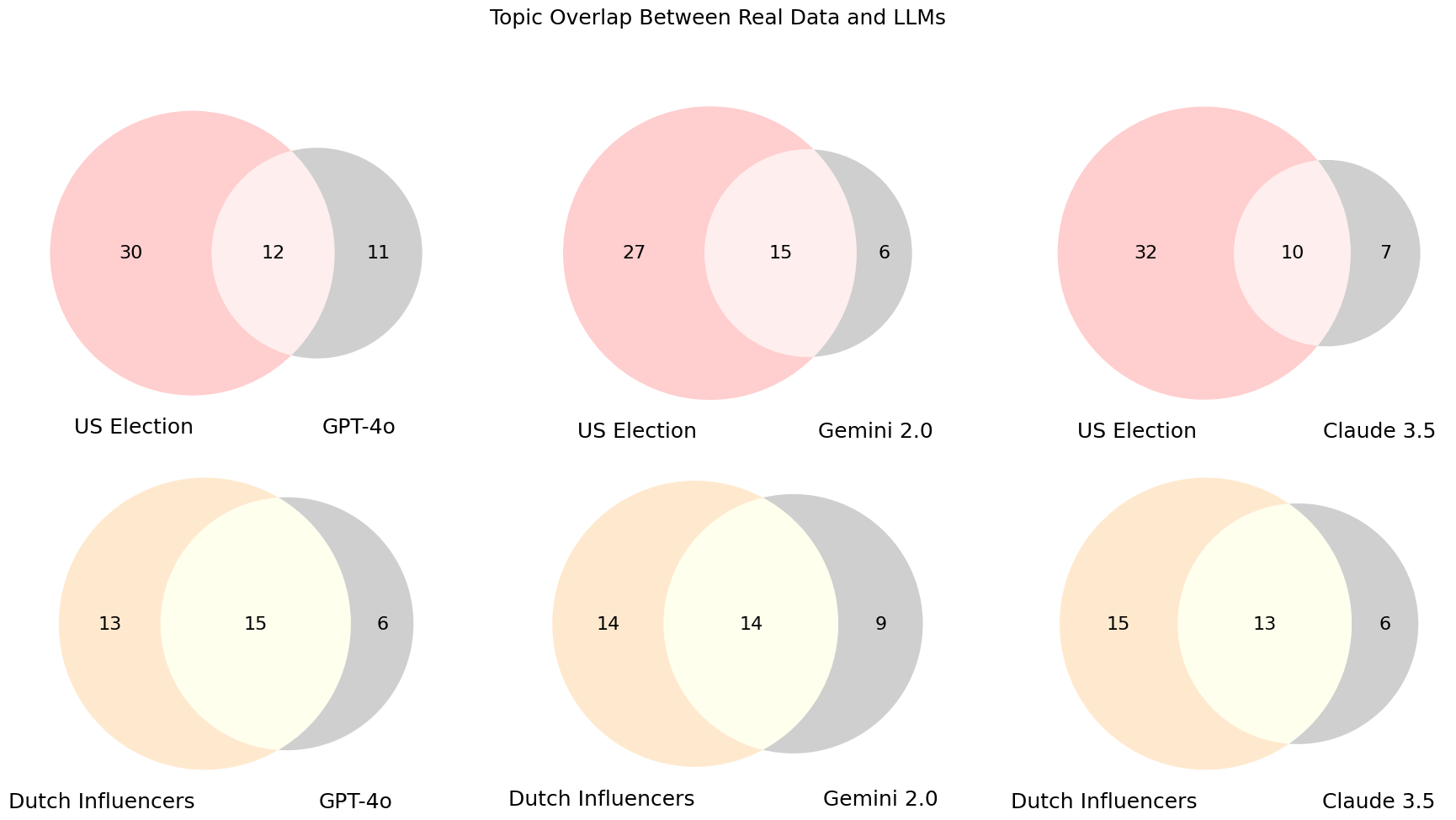}
    \caption{Topic Overlap between real and synthetic data generated using MPTM prompting on the Elections dataset(top row) and Influencers dataset (bottom row). Topic overlap is better in synthetic influencers dataset than election dataset. } \label{fig:topic_overlap_multi_plt}
\end{figure}

\begin{figure}[!htb]
\centering
\begin{subfigure}{0.50\textwidth}
    \centering
    \includegraphics[width=\linewidth, keepaspectratio]{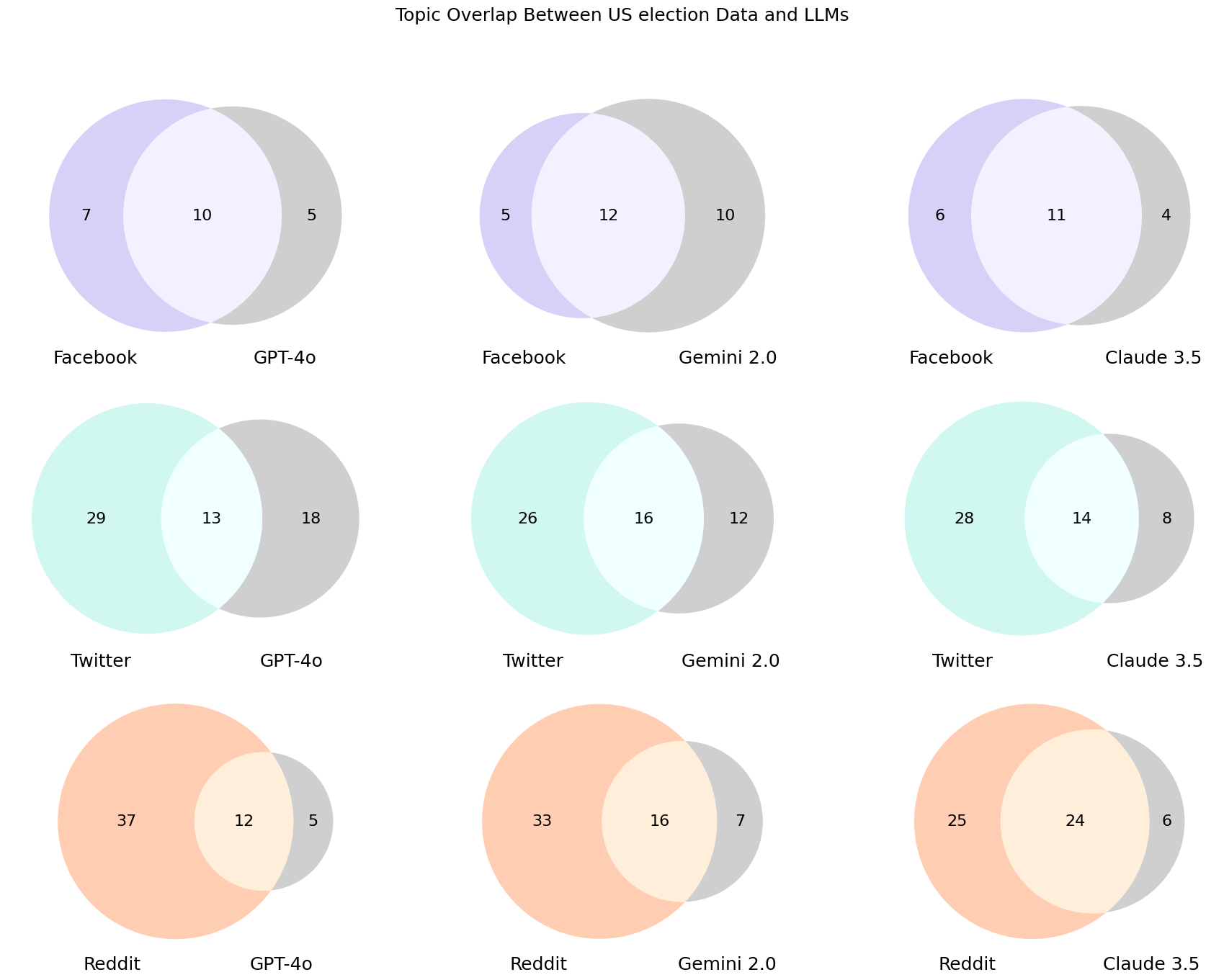}
    \caption{Elections Dataset}
\end{subfigure}\hfill
\begin{subfigure}{0.50\textwidth}
    \centering
    \includegraphics[width=\linewidth, keepaspectratio]{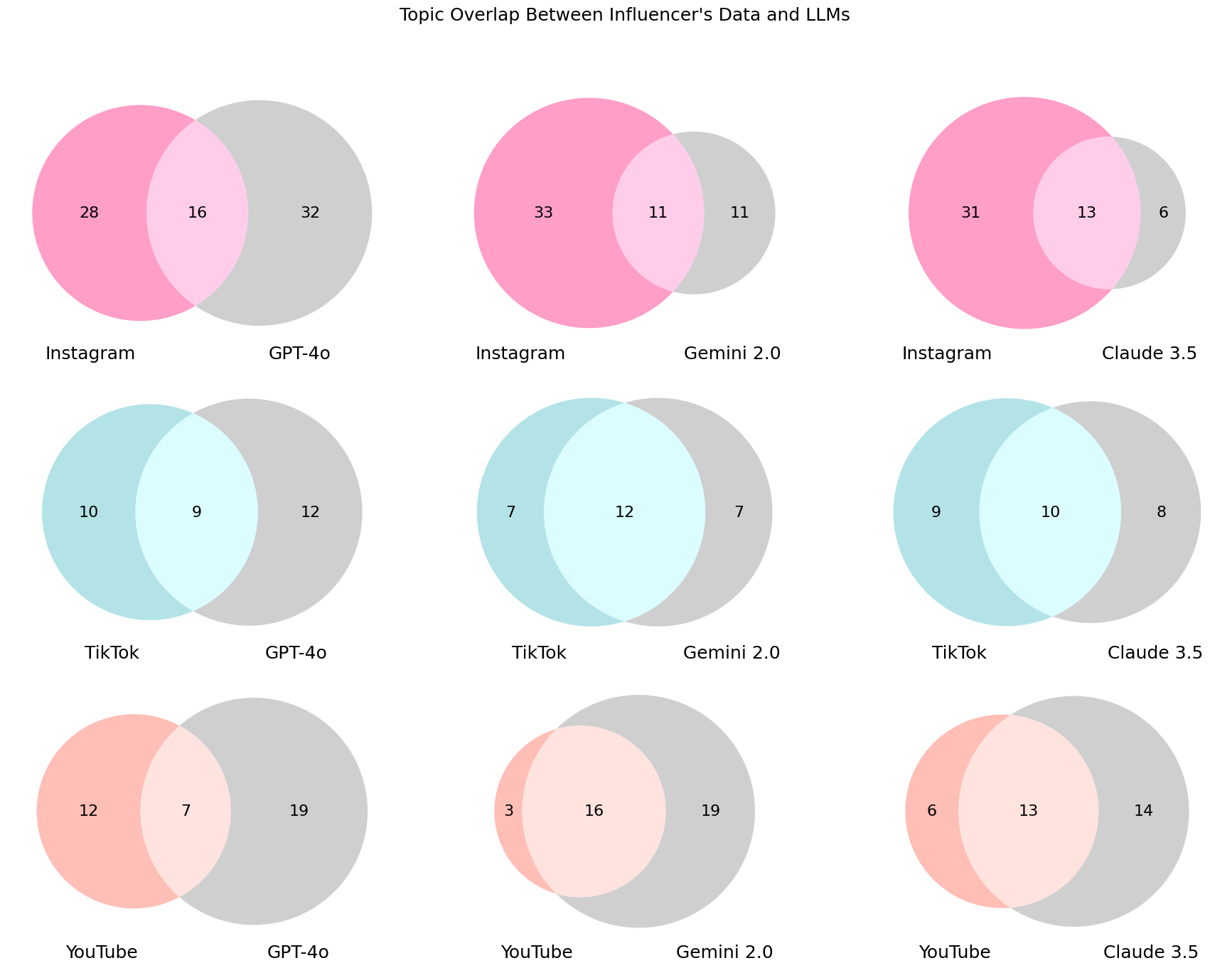}
    \caption{Influencers Dataset}
\end{subfigure}\hfill

\caption{Topic overlap among platforms in per-platform prompting (as proposed by Tari et al.\cite{tari2024leveraging}) between real and synthetic data. (a) US Mid-Term Election Dataset (b) Dutch Influencer's Dataset. Topics on Facebook are better represented in synthetic datasets generated by all LLMs. Gemini-2.0 gives better topic overlap on TikTok and YouTube.}
\label{fig:topic_overlap_per_plt}
\end{figure}

Considering that social media posting behavior revolves around topics, we analyze topics in the synthetic data generated by different LLMs and compare them with topics in real posts. 
Intuitively, the generated data should have topics related to the elections and what influencers choose to talk about. However, new topics in these general subjects are also welcome for the diversity of synthetic data and its potential utility.
For topic extraction, we used BERTopic~\cite{grootendorst2022bertopic}, which is being increasingly used for social media data~\cite{verbeij2024happiness,hanley2023happenstance,hua2022using}.
For generating sentence embeddings required by BERTopic, we used all-MiniLM-L6-v2\footnote{https://huggingface.co/sentence-transformers/all-MiniLM-L6-v2}~\cite{wang2020minilm} sentence transformer that works with sentences as well as short paragraphs, which is useful in our case because of the length variety from long Reddit and YouTube posts to short tweets.
We performed minimal text cleaning for optimization purposes. 
Salient features such as hashtags and emojis were abundant in the influencer dataset, and omitting them resulted in posts losing their meanings. 
Therefore, only URLs and stop words were removed during this process.
We focused only on the topics that appeared in at least 10 social media posts on the corresponding platform.
Figs~\ref{fig:topic_overlap_multi_plt} and~\ref{fig:topic_overlap_per_plt} show the number of common topics across different platforms and unique topics in real and synthetic datasets generated using MPTM prompting and per-platform prompting, respectively. 
To obtain topic overlap, we used topic vectors and computed pairwise cosine similarity, applying an empirically determined threshold of 0.7, and using a greedy algorithm to identify matching topics.  

In MPTM prompting, we observe (Fig~\ref{fig:topic_overlap_multi_plt}) that the elections dataset has more unique topics in real posts than the influencers dataset. The Gemini-2.0 model produces the most common topics in the elections dataset, followed by the ones generated by the GPT-4o and Claude-3.5 models. GPT-4o generated more unique topics, suggesting better capabilities to generate newer topics in the elections dataset than the Gemini-2.0 and Claude-3.5 models. In the case of influencer datasets, all models produce a similar proportion of unique topics in real content and common topics. However, LLM-generated data contains considerably fewer unique topics for all models.
Overall, from the fidelity perspective in the multi-platform settings, this suggests that LLMs can better replicate topics in the influencers dataset rather than the election dataset. In contrast, in the elections dataset, large proportions of topics (30, 27, and 32) are unique in real data but not replicated in synthetic data. 


In per-platform settings (Fig~\ref{fig:topic_overlap_per_plt}), we observe that in the elections dataset, there are more common topics in real and synthetic Facebook content than those in Twitter and Reddit. All LLM models generate the most unique topics on Twitter, suggesting they have better diverse generative capability for tweets.
Reddit has the most unique topics, suggesting that it is challenging for LLMs to replicate this diversity in synthetic data, except for Claude-3.5, which tries to replicate a similar proportion of common topics as unique topics. 
This can be explained by Reddit post lengths being the highest among all platforms, making them the most challenging for the LLMs to replicate.
In the influencers dataset, we find that Instagram has the most unique topics across all LLM-generated data compared to YouTube and TikTok. GPT-4o generates many unique topics across all platforms, suggesting better diversity capabilities. In contrast, Gemini-2.0 and Claude-3.5 exhibit better replicability (more common topics) capabilities for YouTube and TikTok. These models give better diversity capability for only the YouTube platform, where video descriptions are usually longer.

To better understand the actual topic content, Tables~\ref{tab:topics_elections} and~\ref{tab:topics_influencers} present popular topics in real data (Sample Pool) and MPTM prompt-generated synthetic data.
We select the most popular top-5 topic clusters with the most posts. Subsequently, top-3 terms are chosen as the representative words for each topic cluster. 
\begin{table}[!h]
\centering
\caption{Popular topic clusters (top-3 important words per cluster) in real and synthetic elections dataset. Popular topics and themes related to elections are captured in all synthetic datasets.}
\label{tab:topics_elections}
\begin{tabular}{p{2cm}|p{10cm}}
\hline
\textbf{Dataset} & \textbf{Topics (Top 3 Words)} \\
\hline
Sample Pool & 0:[election, ballots, ballot], 1:[abortion, women, republican], 2:[vote republican, republican, vote], 3:[sa, illegitimate, ang], 4:[ukraine, russian, russia] \\
\hline
Claude-3.5 & 0:[today, make, vote], 1:[trump, midterm, political], 2:[media, speech, free speech], 3:[mailin, voting, mailin voting], 4:[diversity, gay, minority] \\
\hline
Gemini-2.0 & 0:[blue, woke, theyre], 1:[election, vote, voting], 2:[community, local, support], 3:[inflation, biden, bidenflation], 4:[midterms, midterm, midterms2024] \\
\hline
GPT-4o & 0:[mailin, ballot, absentee], 1:[medicare, social security, trump], 2:[musk, media, elon], 3:[latino, hispanic, conservative], 4:[midterm elections, midterm, midterms] \\
\hline
\end{tabular}
\end{table}
In the election dataset, all LLMs could match reasonably well in election discourse in general. When we look at specifics, we find none of the LLMs could match popular topic clusters related to Ukraine and Russia. Continuing in this trend, we observe that while real content has women and abortion as popular topics, other LLMS picks up other social groups like \textit{Latino-Hispanic} and \textit{Gay} by GPT-4o and Claude-3.5 models, respectively. Given that we focus exclusively on popular topics, these themes are present across all datasets but in less popular proportions. 
\begin{table}[!h]
\centering
\caption{Popular topic clusters (top-3 important words per cluster) in real and synthetic influencers dataset. Synthetic datasets struggle to capture all popular topics in real data.}
\label{tab:topics_influencers}
\begin{tabular}{p{2cm}|p{10cm}}
\textbf{Dataset} & \textbf{Topics (Top 3 Words)} \\
\hline
Sample Pool & 0:[mijn, en, de], 1:[fashionhacks, hit daily, stylinghacks], 2:[creed, ac, odyssey], 3:[workout, physiotherapy, physio], 4:[makeup, eye makeup, eye] \\
\hline
Claude-3.5 & 0:[xbox, gaming, vintage], 1:[themepark, fy fyp, disney], 2:[horror, minecraft, roblox], 3:[cold, breath, xx], 4:[iamsarkis, de, niet] \\
\hline
Gemini-2.0 & 0:[makeup, hair, products], 1:[recipe, coffee, pasta], 2:[phasmophobia, twitter, link], 3:[workout, gym, fitness], 4:[de, link, op] \\
\hline
GPT-4o & 0:[coaster, whos, universal], 1:[palette, hair, foundation], 2:[recipe, recipes, homemade], 3:[ice, wimhofmethod, cold], 4:[workout, gym, exercise] \\
\hline
\end{tabular}
\end{table} 
In the influencers dataset, we find fashion, makeup, and fitness to be popular themes in real data. GPT-4o and Gemini-2.0 pick up similar topics related to fashion and fitness. Claude-3.5 focuses on gaming and entertainment-related topics and could not pick fashion or fitness among popular themes. Compared to the election dataset, which is focused on the single theme of elections, the influencers dataset has multiple themes, and we can find that LLMs struggle to capture common popular topics in synthetic datasets. Nevertheless, they present popular themes that influencers are typically expected to popularize. 

\begin{figure}[!htb]
\centering
\begin{subfigure}{0.45\textwidth}
    \centering
    \includegraphics[width=\linewidth, keepaspectratio]{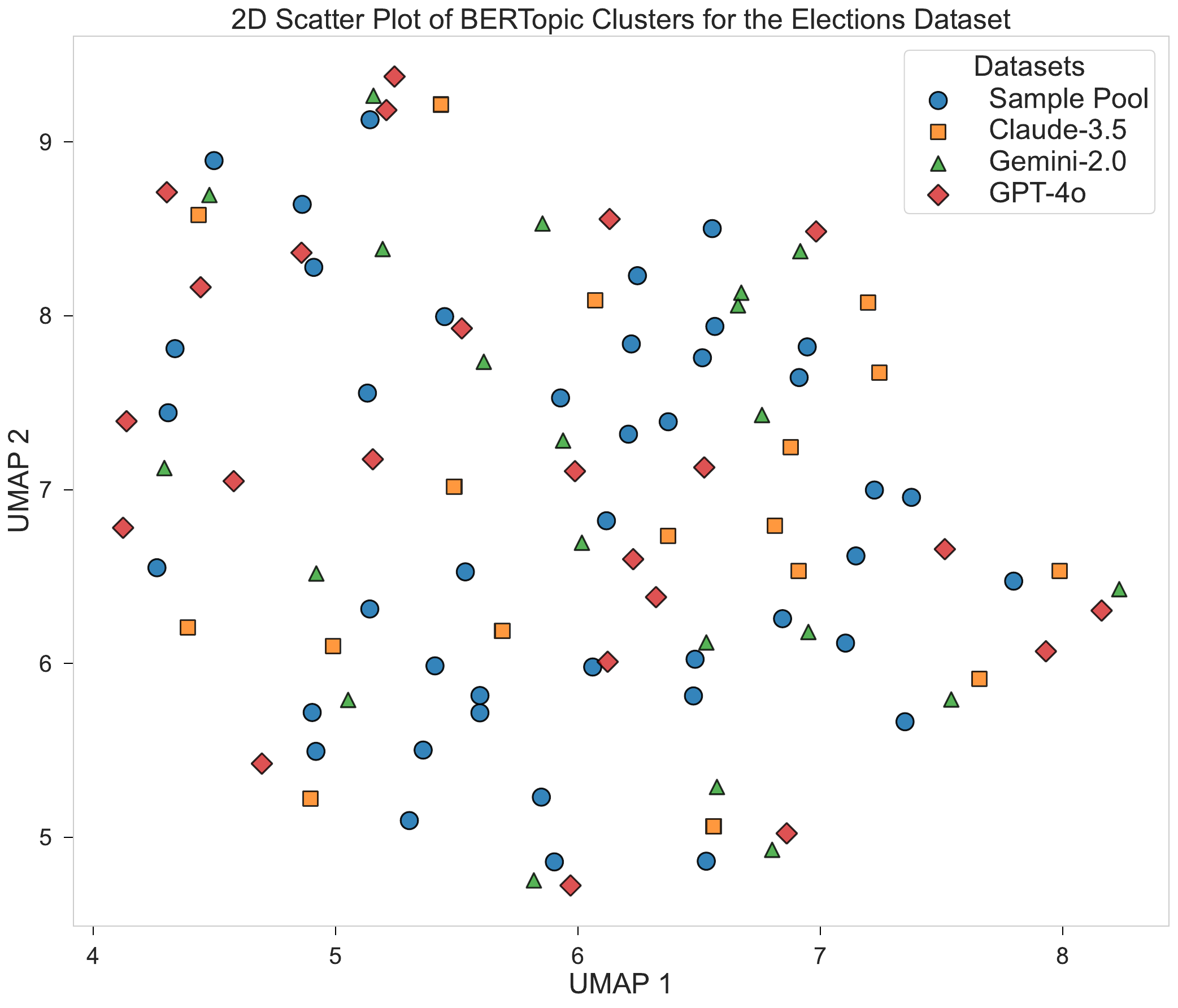}
    \caption{Elections Dataset}
\end{subfigure}
\begin{subfigure}{0.45\textwidth}
    \centering
    \includegraphics[width=\linewidth, keepaspectratio]{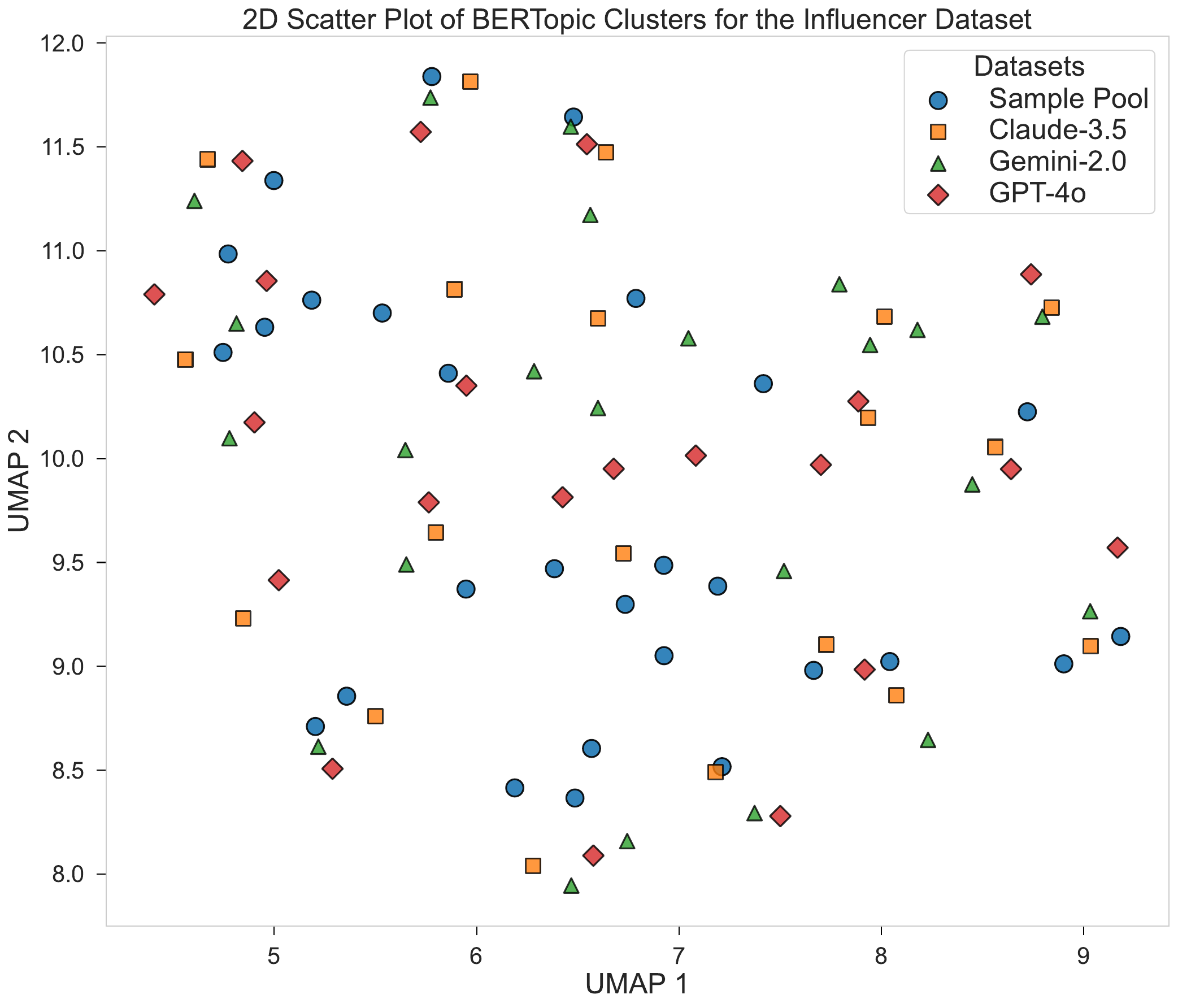}
    \caption{Influencers Dataset}
\end{subfigure}
\caption{Comparison of topics generated by different LLMs using MPTM prompting using dimensionally reduced topic embedding for (a) US Election Dataset (b) Dutch Influencer's Dataset.}
\label{fig:topic_space_compare}
\end{figure}
\begin{table}[!h]
\centering
\small
\caption{Euclidean distances of topic embeddings shown in Fig~\ref{fig:topic_space_compare} for topics in synthetic dataset with respect to the real dataset. Distances are more in influencers dataset for all LLM generated synthetic data.}
\label{tab:topic_space_distance}
\begin{tabular}{lcccc}
\toprule
\textbf{Elections} & \textbf{Avg Distance} & \textbf{Std Dev} & \textbf{Min Distance} & \textbf{Max Distance} \\
\midrule
Claude-3.5  & 1.93 & 0.94 & 0.20 & 4.49 \\
Gemini-2.0  & 1.98 & 0.97 & 0.11 & 4.61 \\
GPT-4o      & 2.06 & 0.98 & 0.07 & 4.69 \\
\bottomrule
\end{tabular}
\begin{tabular}{lcccc}
\toprule
\textbf{Influencer} & \textbf{Avg Distance} & \textbf{Std Dev} & \textbf{Min Distance} & \textbf{Max Distance} \\
\midrule
Claude-3.5  & 2.13 & 1.03 & 0.04 & 5.06 \\
Gemini-2.0  & 2.12 & 1.00 & 0.05 & 5.03 \\
GPT-4o      & 2.09 & 1.04 & 0.11 & 5.05 \\
\bottomrule
\end{tabular}
\end{table}
Figure~\ref{fig:topic_space_compare} presents the topic embeddings of different topic clusters obtained by the BERTopic model for real and synthetic data generated by different LLMs, represented using UMAP dimensionality reduction. By comparing embeddings in topic space, we can find the closeness of topics in synthetic data compared to the real data. Through visual inspection, we find that topic embedding vectors for the elections datasets are closer in the reduced two-dimensional topic space, suggesting that LLMs are better at capturing topics of the real datasets in a multi-platform setting.
Table~\ref{tab:topic_space_distance} presents the Euclidean distances between topic embeddings of the synthetic and real datasets. The results align with the trends observed in Tables~\ref{tab:topics_elections} and~\ref{tab:topics_influencers}, showing that LLMs capture election-related topics more effectively than influencer-related ones, as indicated by lower values of average distances, standard deviations, and min-max ranges.

\begin{table}[!h]
\centering
\caption{Detailed similarity statistics across platforms and datasets for each LLM. Includes average similarity, count of highly similar responses (similarity threshold $>$ 0.8), and fairly similar responses (similarity threshold between 0.6 and 0.8).}
\footnotesize
\begin{tabular}{@{}llccccc@{}}
\toprule
\multicolumn{6}{c}{\textbf{US Election Dataset}} \\
\hline
& & & \multicolumn{2}{c}{\textbf{Recall Count}} & \\
\textbf{Platform} & \textbf{LLM} & \textbf{Avg. Sim.} & \textbf{$>0.8$} & \textbf{$0.6-0.8$} & \textbf{Real Count} \\
\midrule
MPTM     & Claude-3.5 & 0.2306 & 0  & 109 & 999 \\
         & Gemini-2.0 & 0.2176 & 2  & 131 & 999 \\
         & GPT-4o     & 0.2311 & 7  & 165 & 999 \\
Facebook & Claude-3.5 & 0.2433 & 20 & 287 & 734 \\
         & Gemini-2.0 & 0.2314 & 57 & 270 & 734 \\
         & GPT-4o     & 0.2590 & 34 & 274 & 734 \\
Twitter  & Claude-3.5 & 0.2322 & 3  & 145 & 968 \\
         & Gemini-2.0 & 0.2271 & 21 & 164 & 968 \\
         & GPT-4o     & 0.2367 & 1  & 109 & 968 \\
Reddit   & Claude-3.5 & 0.2649 & 1  & 202 & 995 \\
         & Gemini-2.0 & 0.2515 & 30 & 246 & 995 \\
         & GPT-4o     & 0.2590 & 6  & 174 & 995 \\
\midrule
\multicolumn{6}{c}{\textbf{Dutch Influencers Dataset}} \\
\hline
& & & \multicolumn{2}{c}{\textbf{Recall Count}} & \\
\textbf{Platform} & \textbf{LLM} & \textbf{Avg. Sim.} & \textbf{$>0.8$} & \textbf{$0.6-0.8$} & \textbf{Real Count} \\
\midrule
MPTM      & Claude-3.5 & 0.1988 & 0   & 190 & 999 \\
          & Gemini-2.0 & 0.1784 & 9   & 237 & 999 \\
          & GPT-4o     & 0.2042 & 4   & 137 & 999 \\
Instagram & Claude-3.5 & 0.2469 & 8   & 226 & 991 \\
          & Gemini-2.0 & 0.2284 & 21  & 256 & 991 \\
          & GPT-4o     & 0.2475 & 1   & 166 & 991 \\
TikTok    & Claude-3.5 & 0.2498 & 12  & 267 & 977 \\
          & Gemini-2.0 & 0.2411 & 31  & 300 & 977 \\
          & GPT-4o     & 0.2439 & 0   & 114 & 977 \\
YouTube   & Claude-3.5 & 0.3827 & 52  & 752 & 989 \\
          & Gemini-2.0 & 0.3661 & 292 & 558 & 989 \\
          & GPT-4o     & 0.3947 & 92  & 683 & 989 \\
\bottomrule
\end{tabular}
\label{tab:embedding_space_similarity_results}
\end{table}
\subsection{Embedding Similarity}
Most downstream tasks are solved by converting text into embedding vectors. Therefore, assessing the similarity of generated synthetic content with real content in embedding space is important.
To evaluate the similarity between real and synthetic social media posts, we use OpenAI's \textit{text-embedding-3-large} model~\cite{neelakantan2022text} to convert all text into embedding vectors. We then compute pairwise cosine similarity for all possible pairs drawn from real and synthetic posts. Table~\ref{tab:embedding_space_similarity_results} presents the average similarity. 
In MPTM prompting, we find that synthetic influencers dataset posts are comparatively more similar to their real counterpart than those of the election dataset.
In the case of long posts (Reddit and YouTube), the embedding similarity is relatively larger than we would expect. 
Among LLMs, GPT-4o-generated posts are marginally more similar, whereas Gemini-2.0-generated posts are slightly farther apart.
However, the differences are too small to matter much.
The average similarity of embeddings confirms the findings of Tari et al.\cite{tari2024leveraging}.

\begin{figure}[H]
    \centering
    \begin{subfigure}[t]{0.32\textwidth}
        \includegraphics[width=\linewidth]{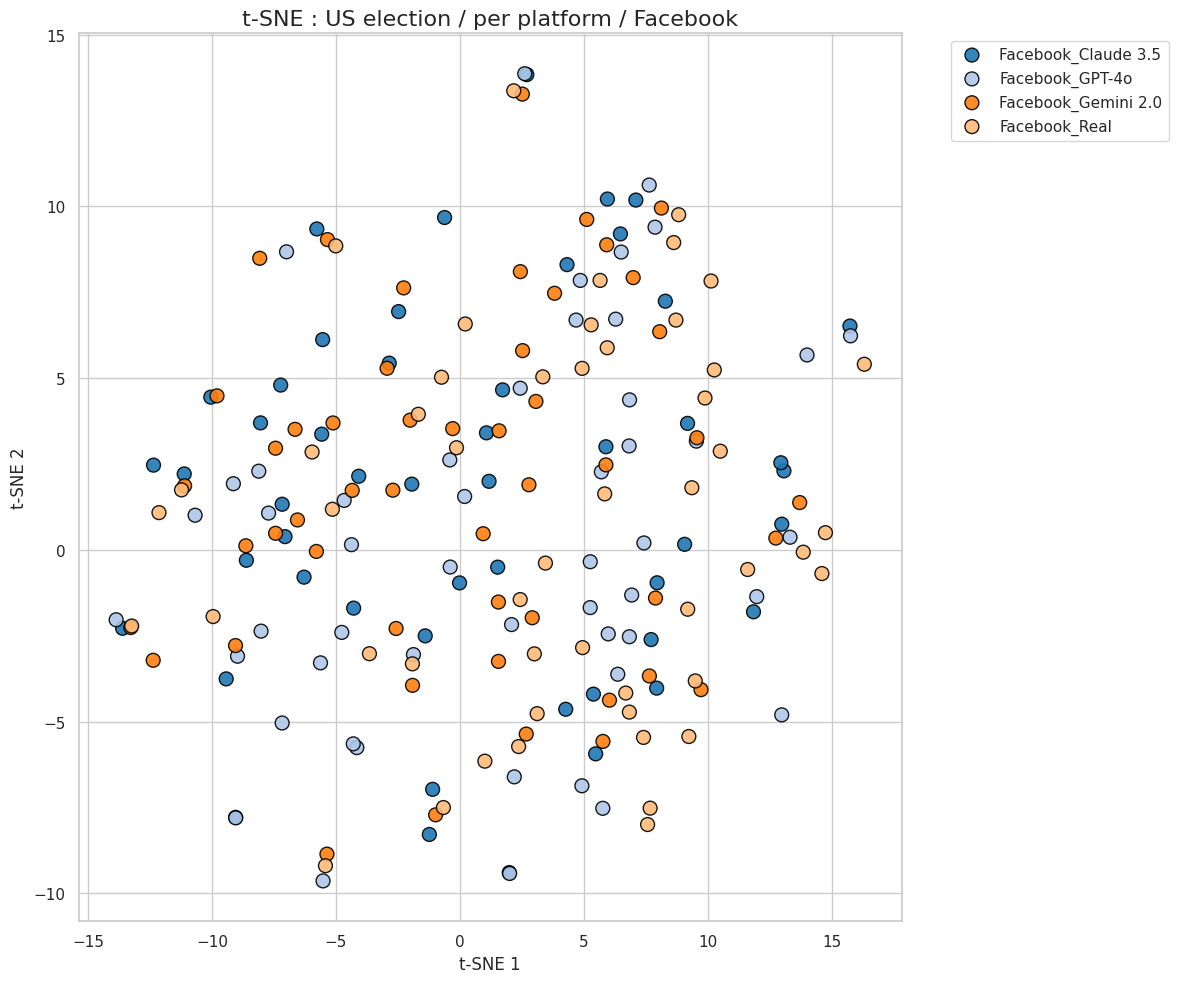}
        \caption{US – Facebook}
    \end{subfigure}
    \begin{subfigure}[t]{0.32\textwidth}
        \includegraphics[width=\linewidth]{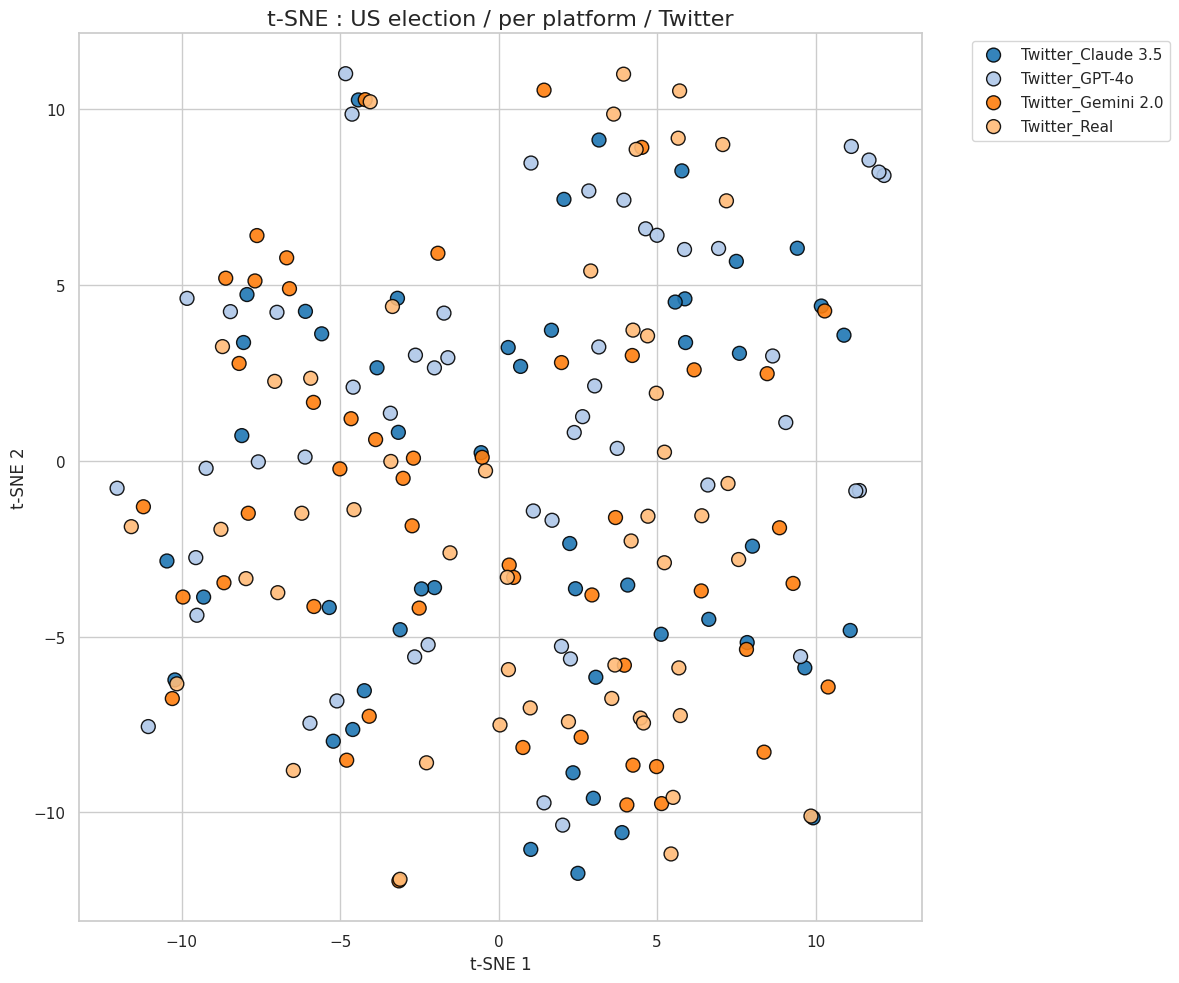}
        \caption{US – Twitter}
    \end{subfigure}
    \begin{subfigure}[t]{0.32\textwidth}
        \includegraphics[width=\linewidth]{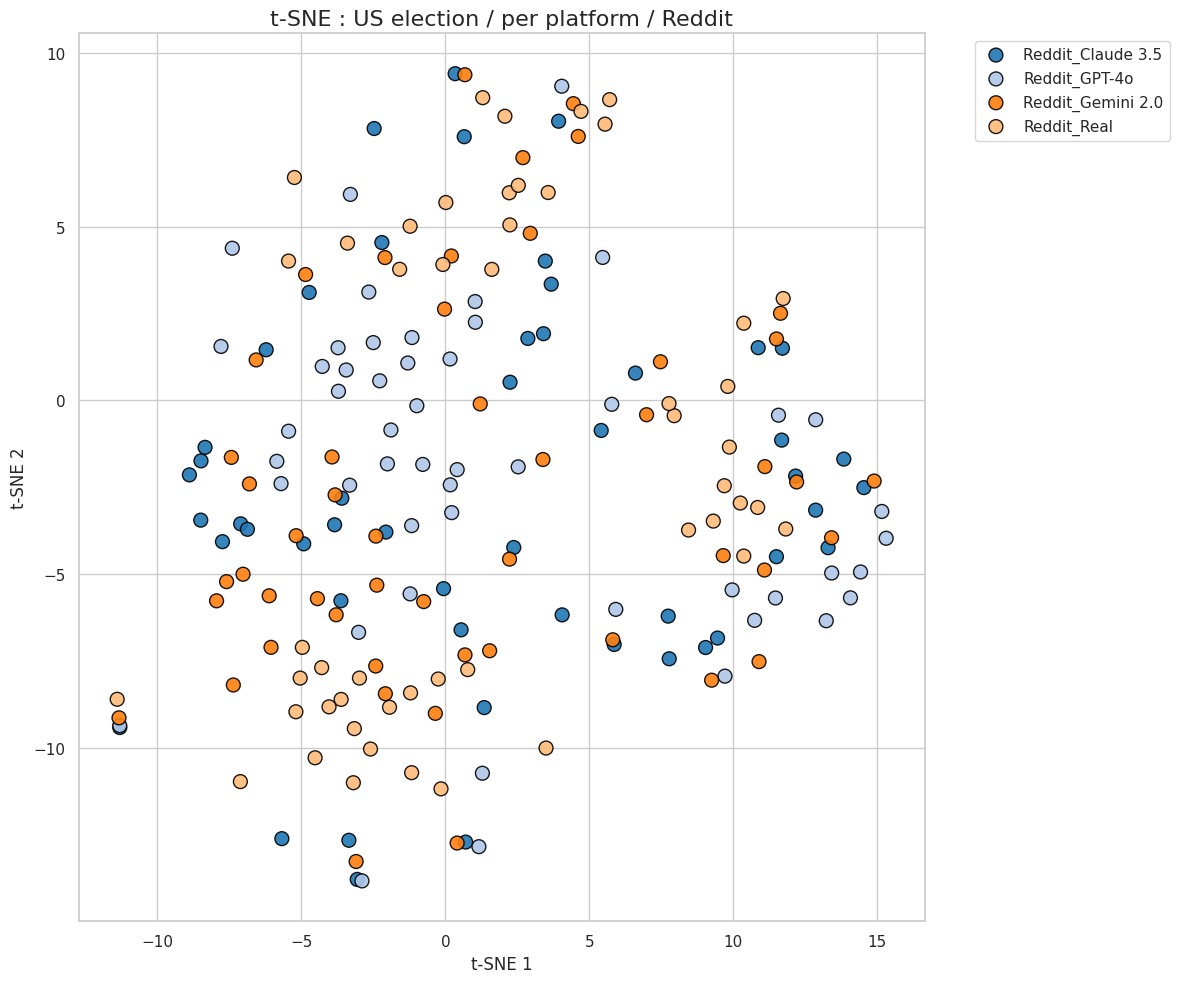}
        \caption{US – Reddit}
    \end{subfigure}

    \begin{subfigure}[t]{0.32\textwidth}
        \includegraphics[width=\linewidth]{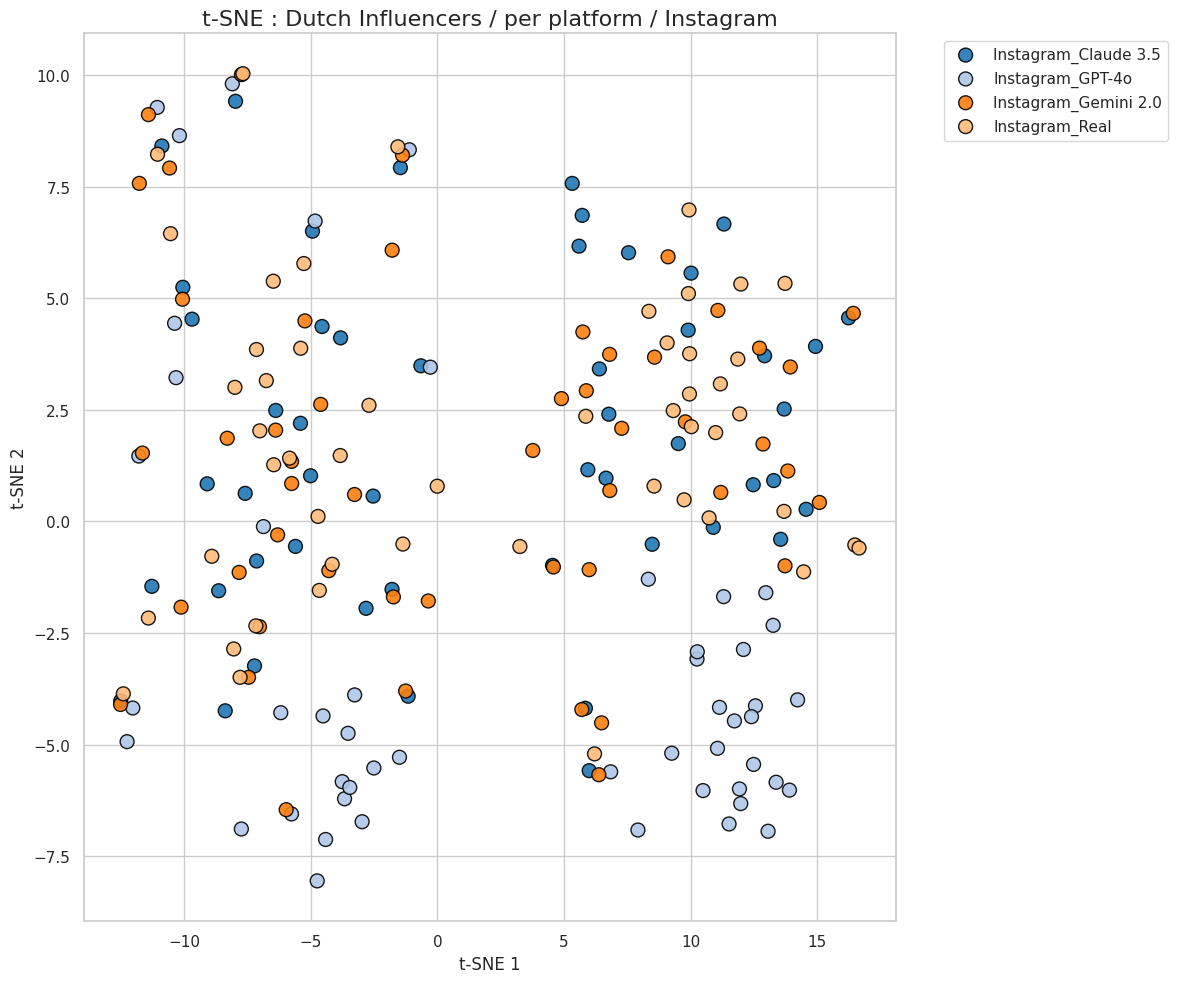}
        \caption{Dutch – Instagram}
    \end{subfigure}
    \begin{subfigure}[t]{0.32\textwidth}
        \includegraphics[width=\linewidth]{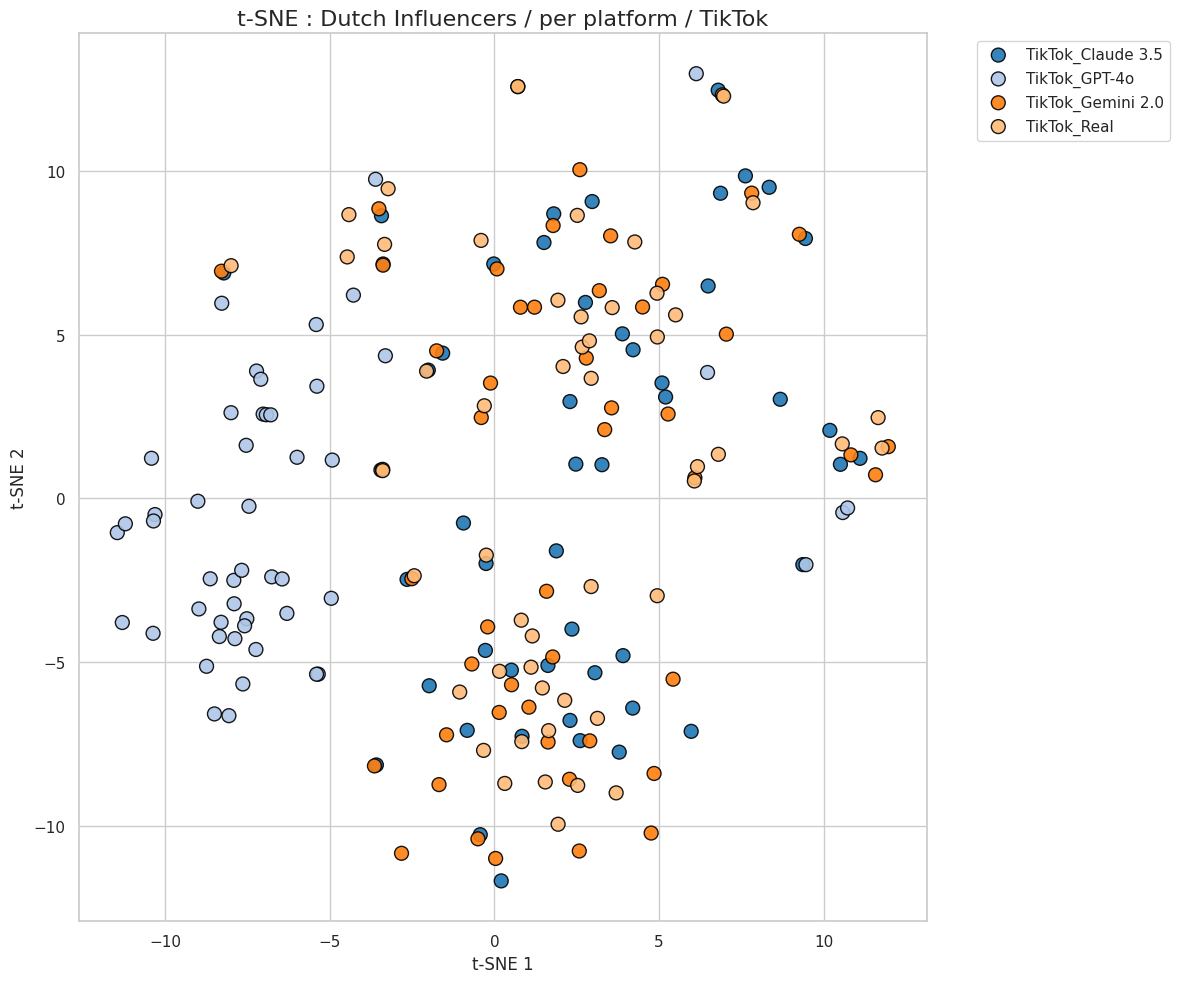}
        \caption{Dutch – TikTok}
    \end{subfigure}
    \begin{subfigure}[t]{0.32\textwidth}
        \includegraphics[width=\linewidth]{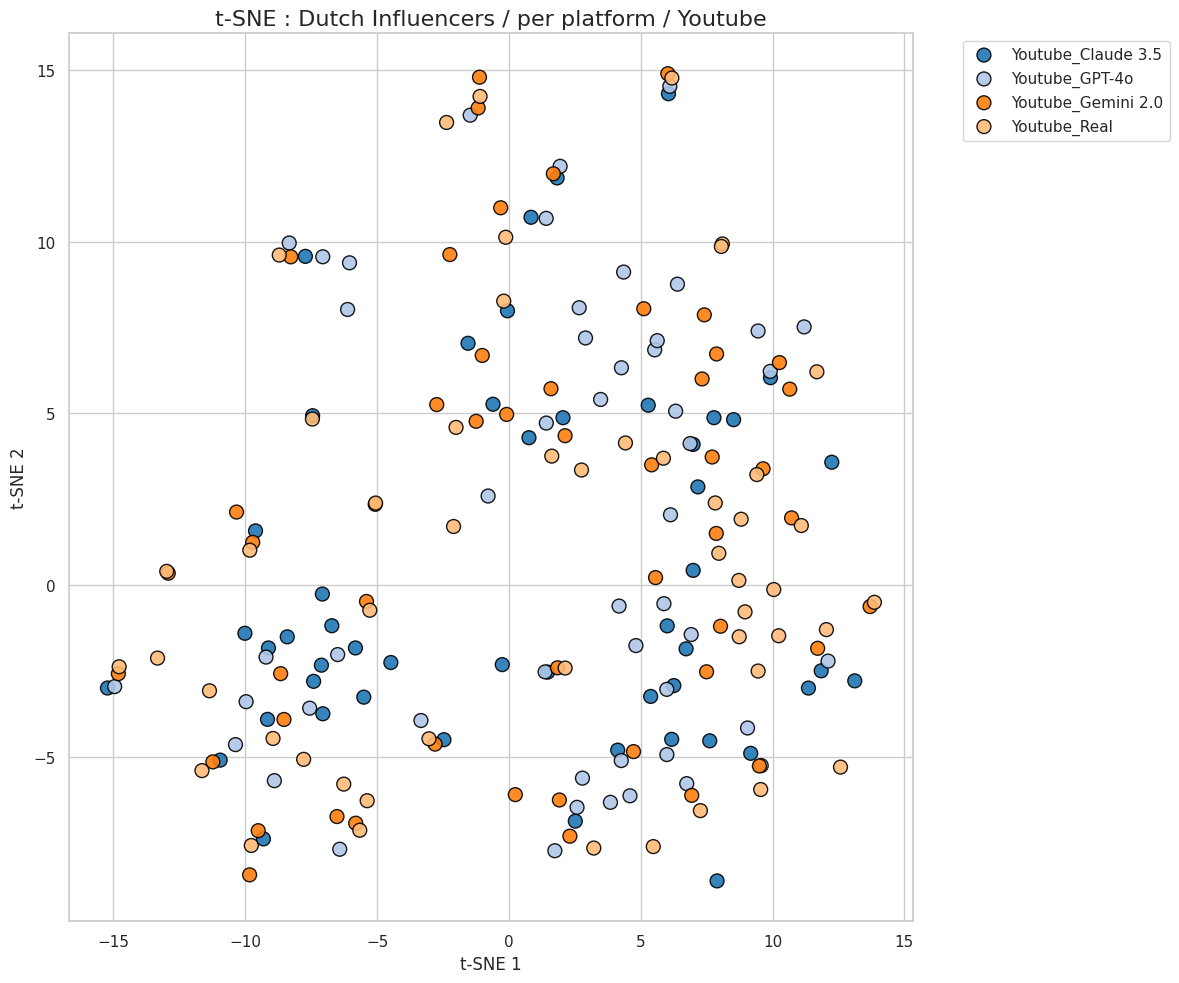}
        \caption{Dutch – YouTube}
    \end{subfigure}

    \begin{subfigure}[t]{0.45\textwidth}
        \includegraphics[width=\linewidth]{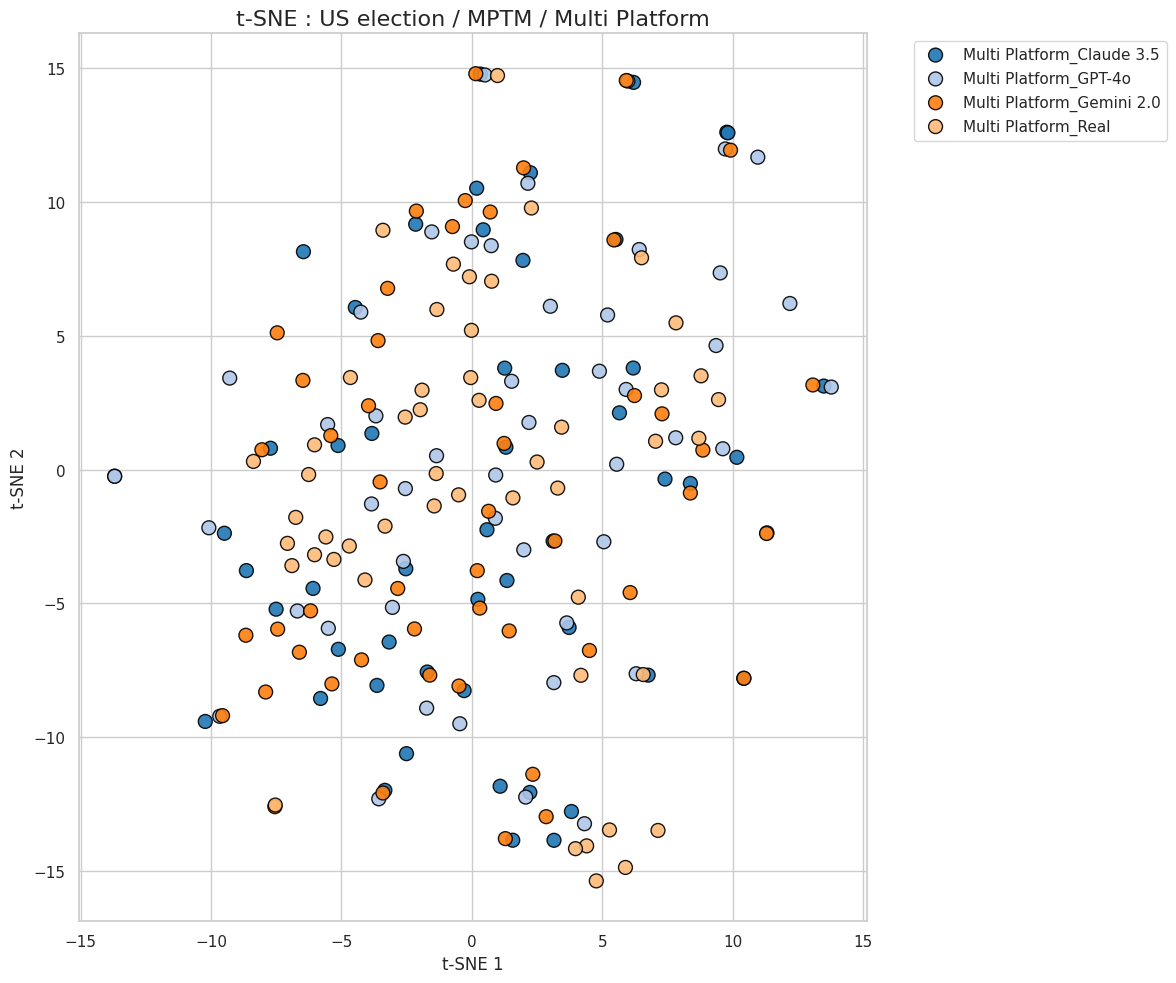}
        \caption{US – Multi-Platform}
    \end{subfigure}
    \begin{subfigure}[t]{0.45\textwidth}
        \includegraphics[width=\linewidth]{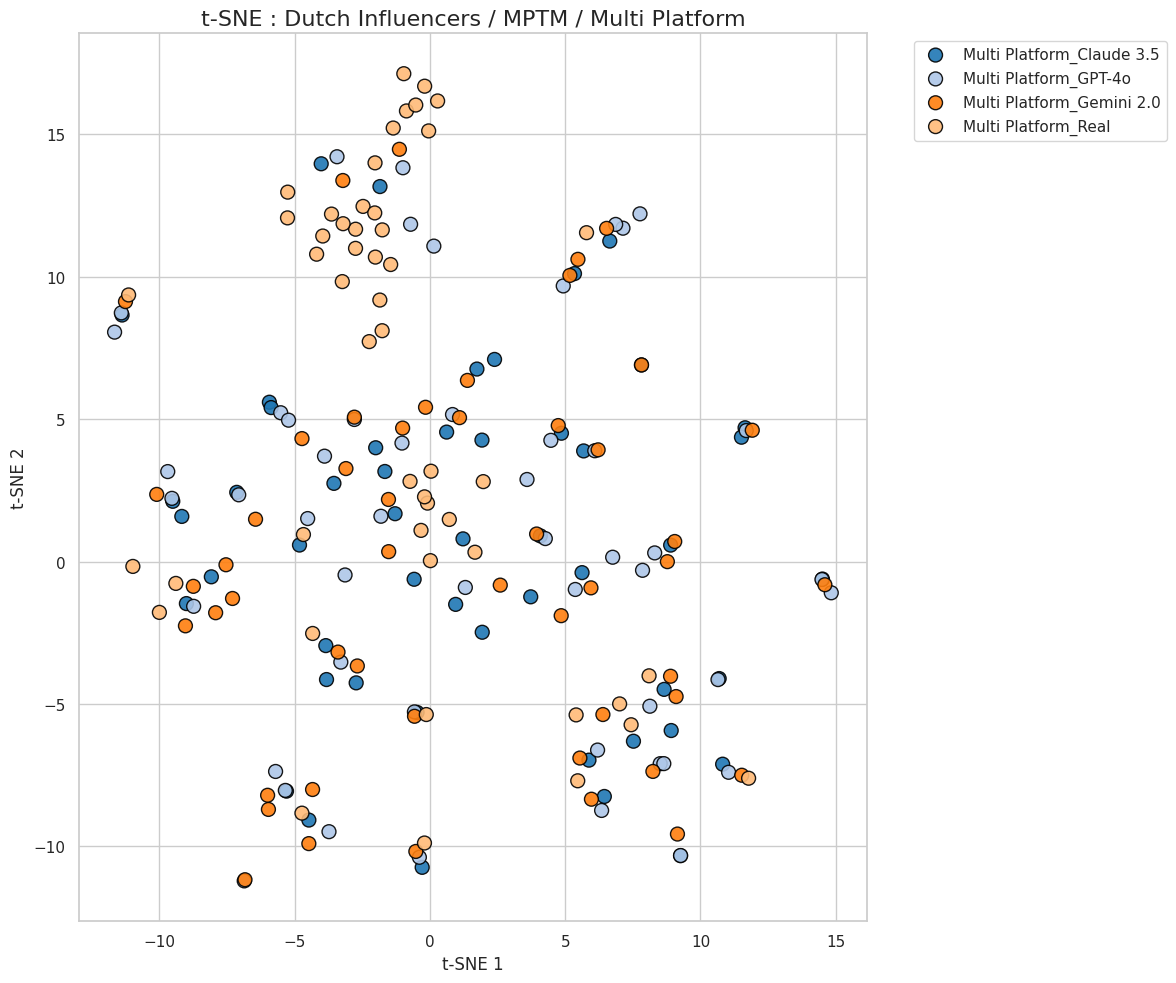}
        \caption{Dutch – Multi-Platform}
    \end{subfigure}

    \caption{t-SNE visualizations of US and Dutch datasets across different social media platforms.}
    \label{fig:tsne_summary}
\end{figure}

Next, we compute the recall count as follows: for each post in the synthetic dataset, we compare it with all posts in the real dataset and report a match if the cosine similarity between them is greater than 0.8 (highly similar) or is between 0.6 and 0.8 (fairly similar). Table~\ref{tab:embedding_space_similarity_results} presents the results.
In MPTM prompting, we find very few synthetic posts that are highly similar in both the elections and influencers datasets. However, when we relax cosine similarity and move towards fairly similar posts, we find more GPT-4o-generated posts similar to real posts than Gemini-2.0 and Claude-3.5 models in the elections dataset. However, in the influencers dataset, Gemini-2.0 generates more fairly similar posts. 
In the per-platform prompting for elections dataset, we observe that posts on Facebook are more highly and fairly similar, followed by Reddit and Twitter.  
In the influencers dataset, all LLMs can generate YouTube posts, which are the most highly and fairly similar to the real posts.
To visualize embedding vectors, we create t-SNE plots (Fig~\ref{fig:tsne_summary}) using the embeddings of both real and synthetic datasets generated by MPTM prompting and per-platform prompting.
To help in visual inspection, we reduce the embedding vector projections by applying K-means clustering (K=50) and representing only the cluster centroids as dots in Fig~\ref{fig:tsne_summary}.
We experimented with different perplexity values and found consistent results; the plots are drawn on the default perplexity value.
While t-SNE has limitations, it presents a good estimate for preserving local structure in data. 
In the election dataset, synthetic data shows high similarity with real data for Facebook. Synthetic data clusters are closely located around synthetically generated data clusters. This is also corroborated by findings in Table~\ref{tab:embedding_space_similarity_results}. 
In the case of Reddit and Twitter, synthetic embedding clusters are not close to their real counterparts.
In the influencer dataset, we observe high similarity between real and synthetic embedding clusters based on video descriptions on the YouTube platform. This suggests that LLMs can replicate YouTube data, as shown in Table~\ref{tab:embedding_space_similarity_results}.
For Instagram and TikTok, we see many GPT-4o-generated embedding clusters that are far apart from other clusters in the embedding space. 
In the case of embedding clusters obtained by MPTM prompt-generated synthetic data, we find more closeness in the elections dataset than the influencers dataset. We observe a large cluster of real data in the influencers dataset located separately from the rest of the clusters. 

\subsection{Named Entity Analysis}
\begin{figure}[H]
    \centering
    \begin{subfigure}[t]{0.24\textwidth}
        \includegraphics[width=\linewidth]{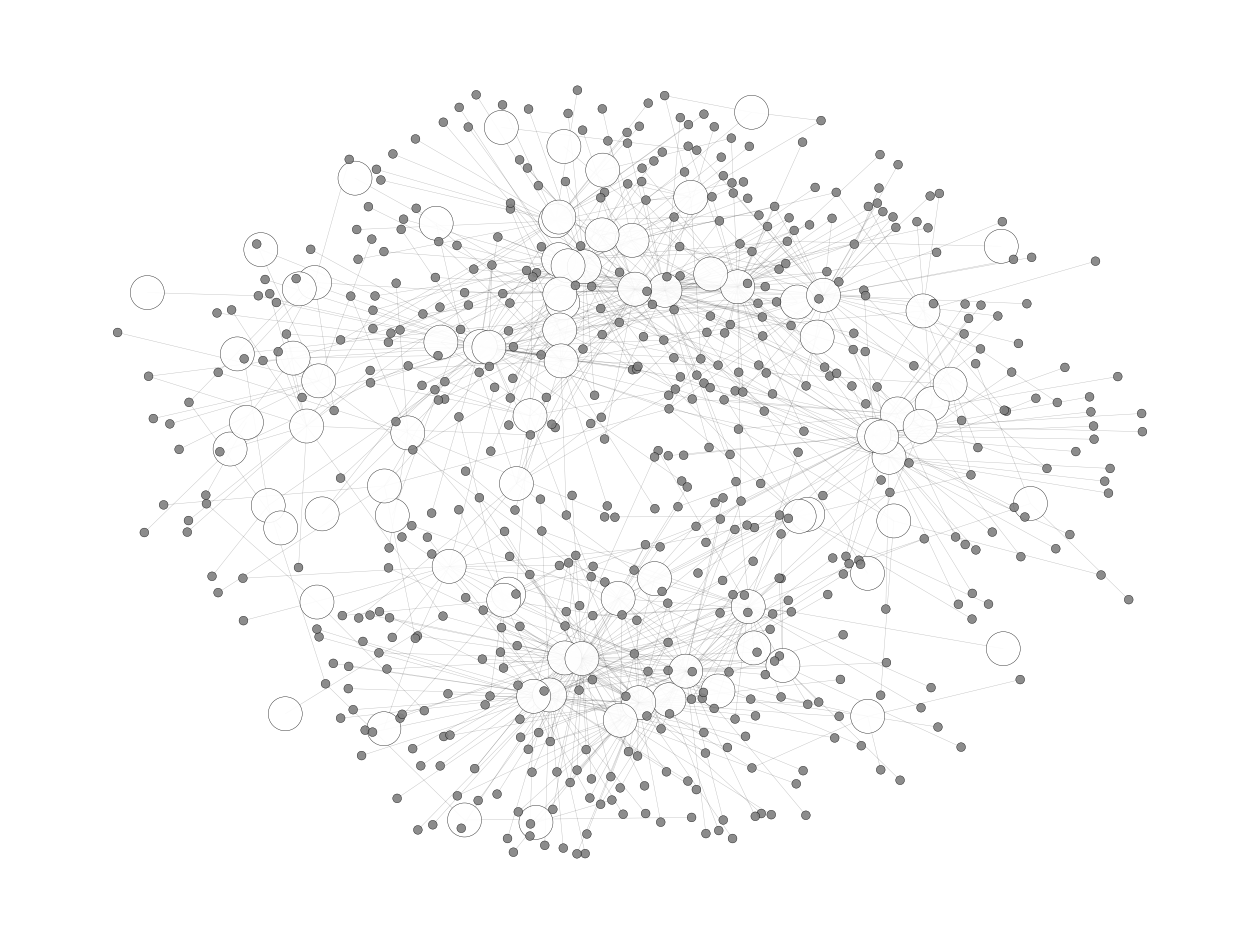}
        \caption{MP Real}
    \end{subfigure}
    \begin{subfigure}[t]{0.24\textwidth}
        \includegraphics[width=\linewidth]{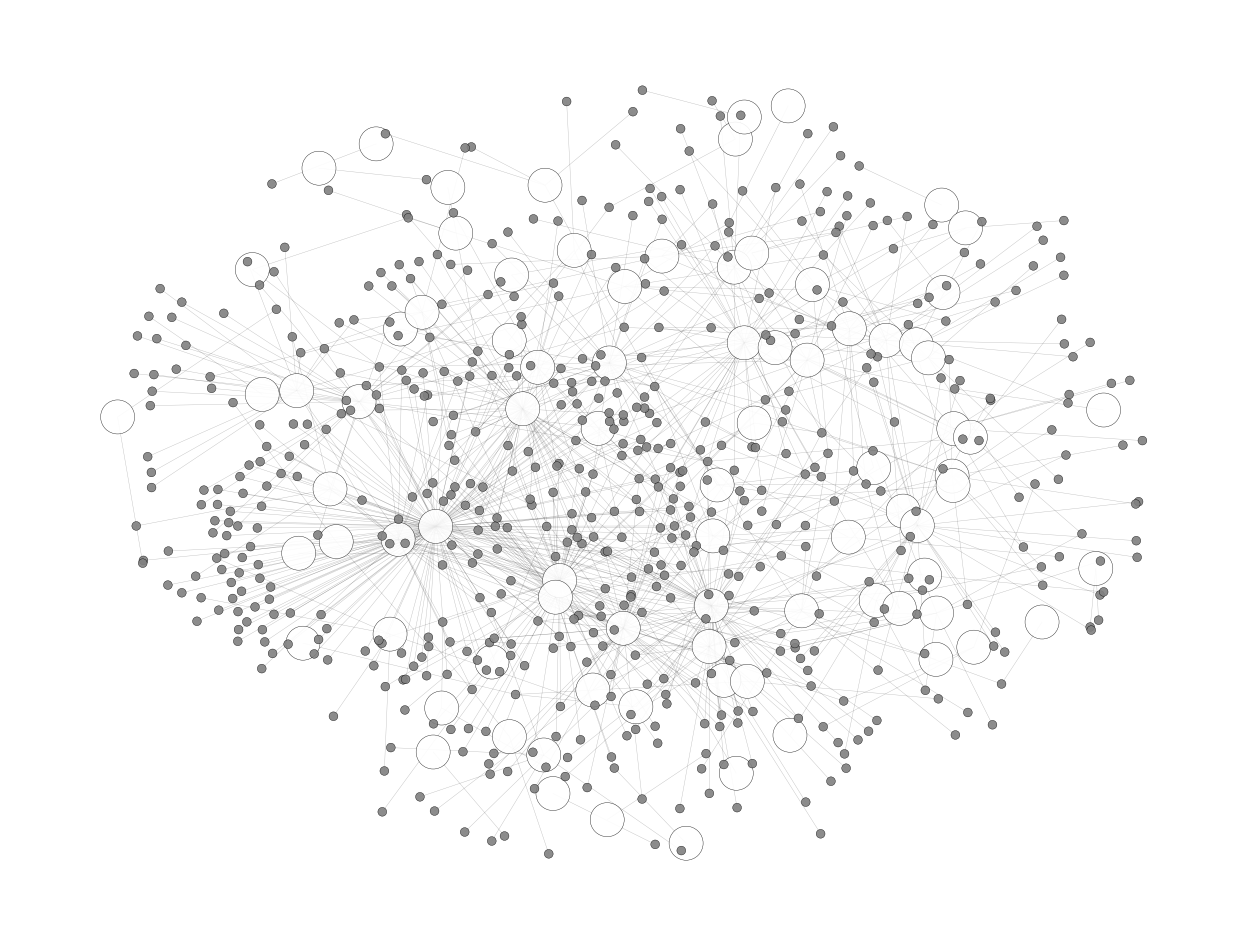}
        \caption{MPTM Claude-3.5}
    \end{subfigure}
    \begin{subfigure}[t]{0.24\textwidth}
        \includegraphics[width=\linewidth]{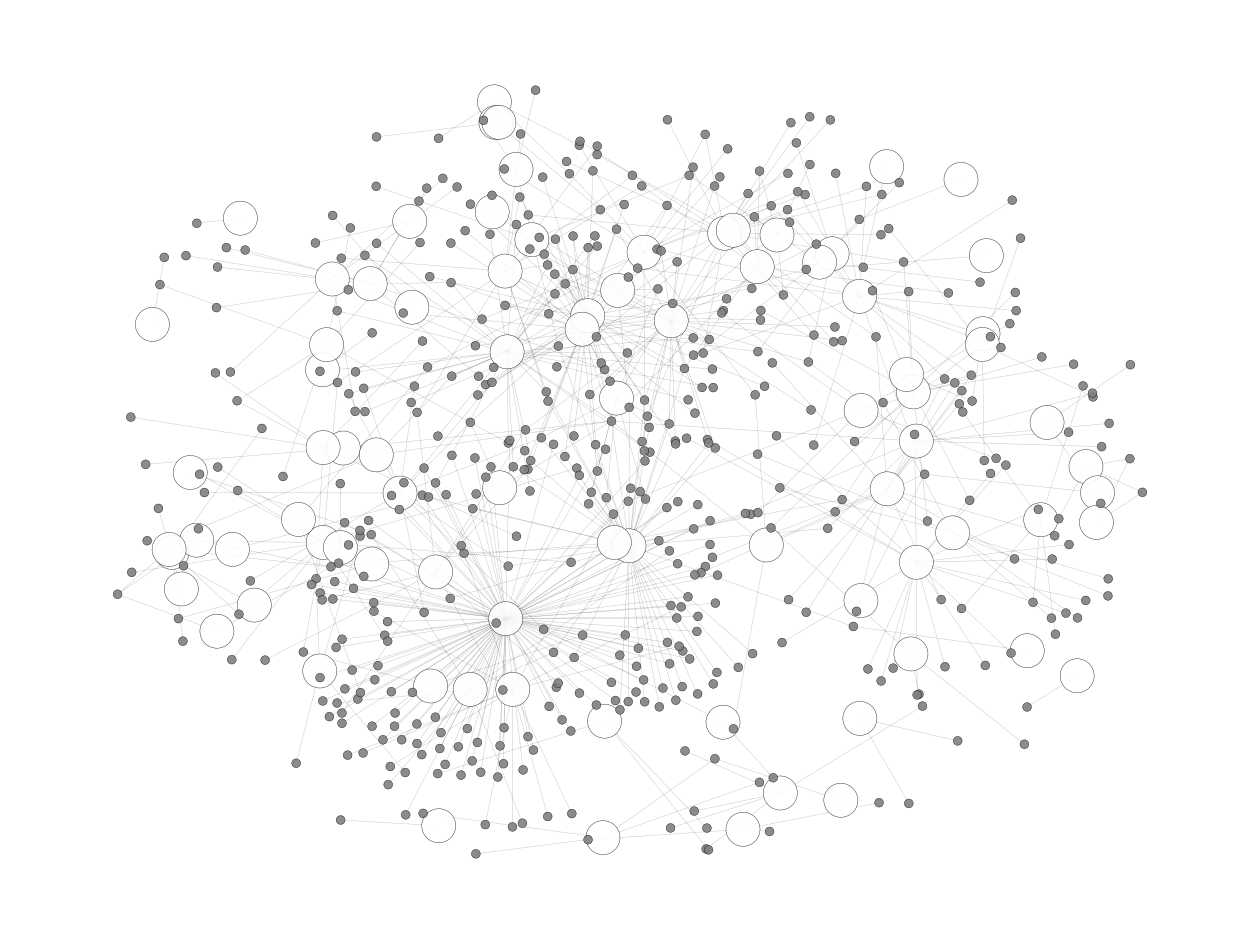}
        \caption{MPTM Gemini-2.0}
    \end{subfigure}
    \begin{subfigure}[t]{0.24\textwidth}
        \includegraphics[width=\linewidth]{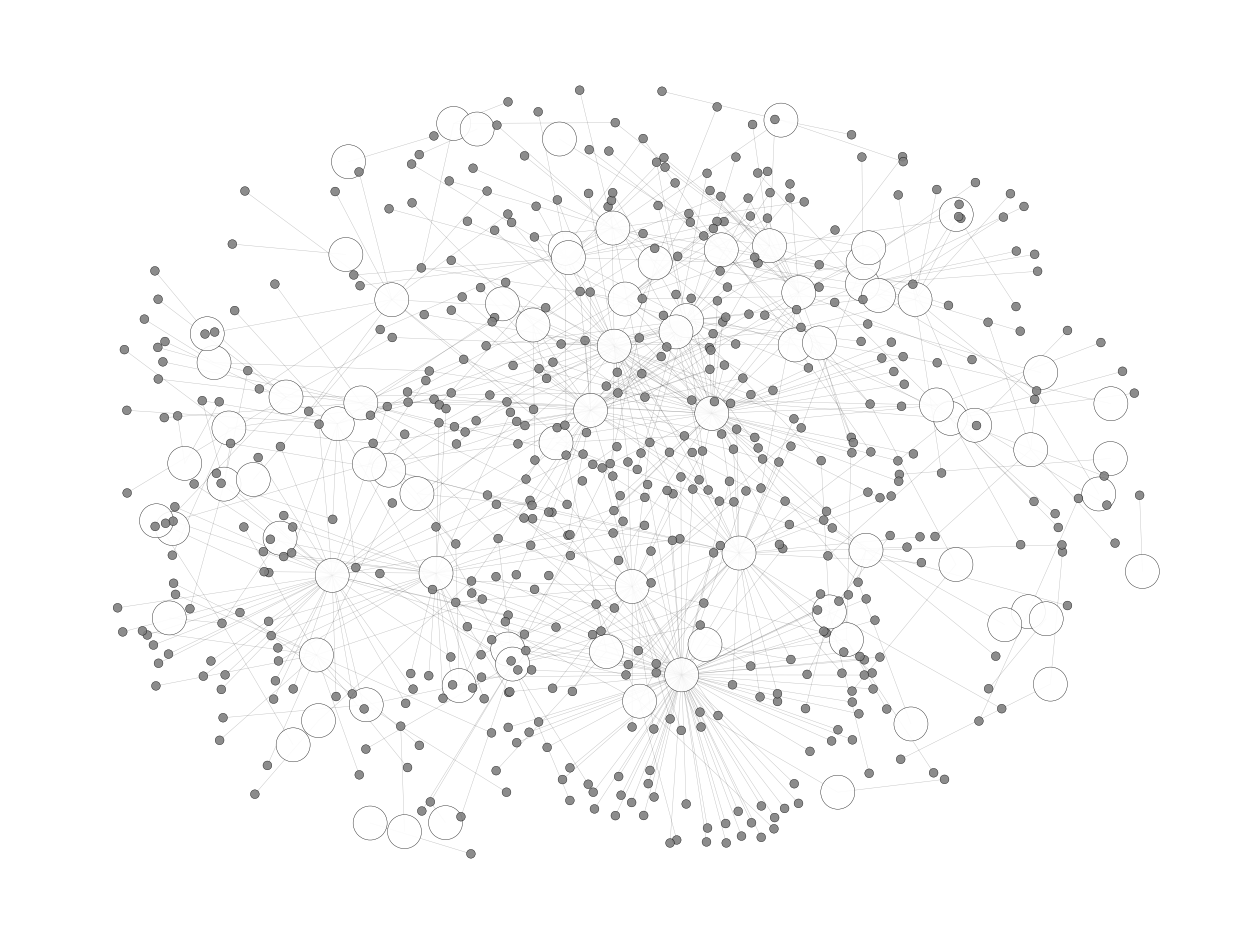}
        \caption{MPTM GPT-4o}
    \end{subfigure}

    \begin{subfigure}[t]{0.24\textwidth}
        \includegraphics[width=\linewidth]{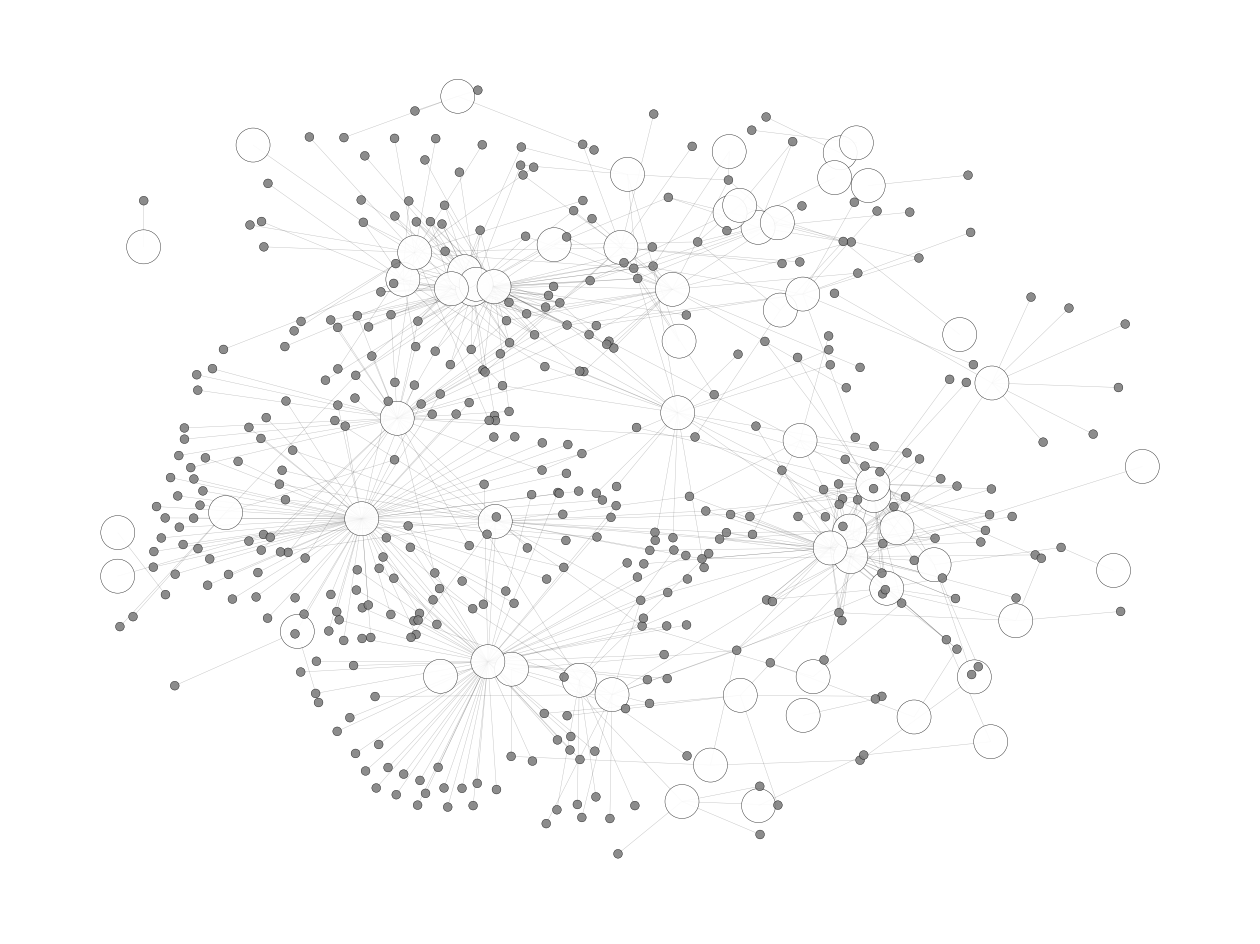}
        \caption{MP Real}
    \end{subfigure}
    \begin{subfigure}[t]{0.24\textwidth}
        \includegraphics[width=\linewidth]{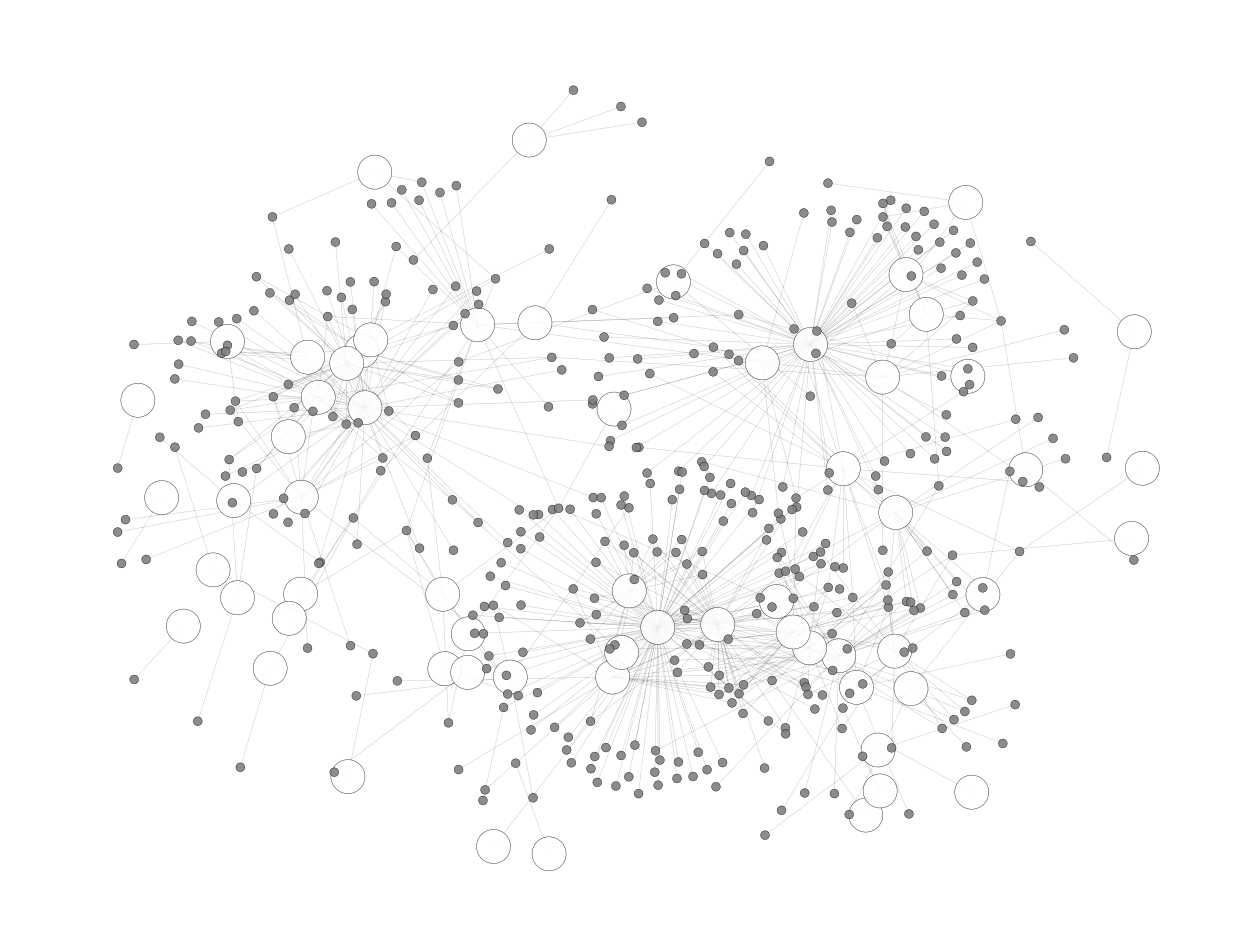}
        \caption{MPTM Claude-3.5}
    \end{subfigure}
    \begin{subfigure}[t]{0.24\textwidth}
        \includegraphics[width=\linewidth]{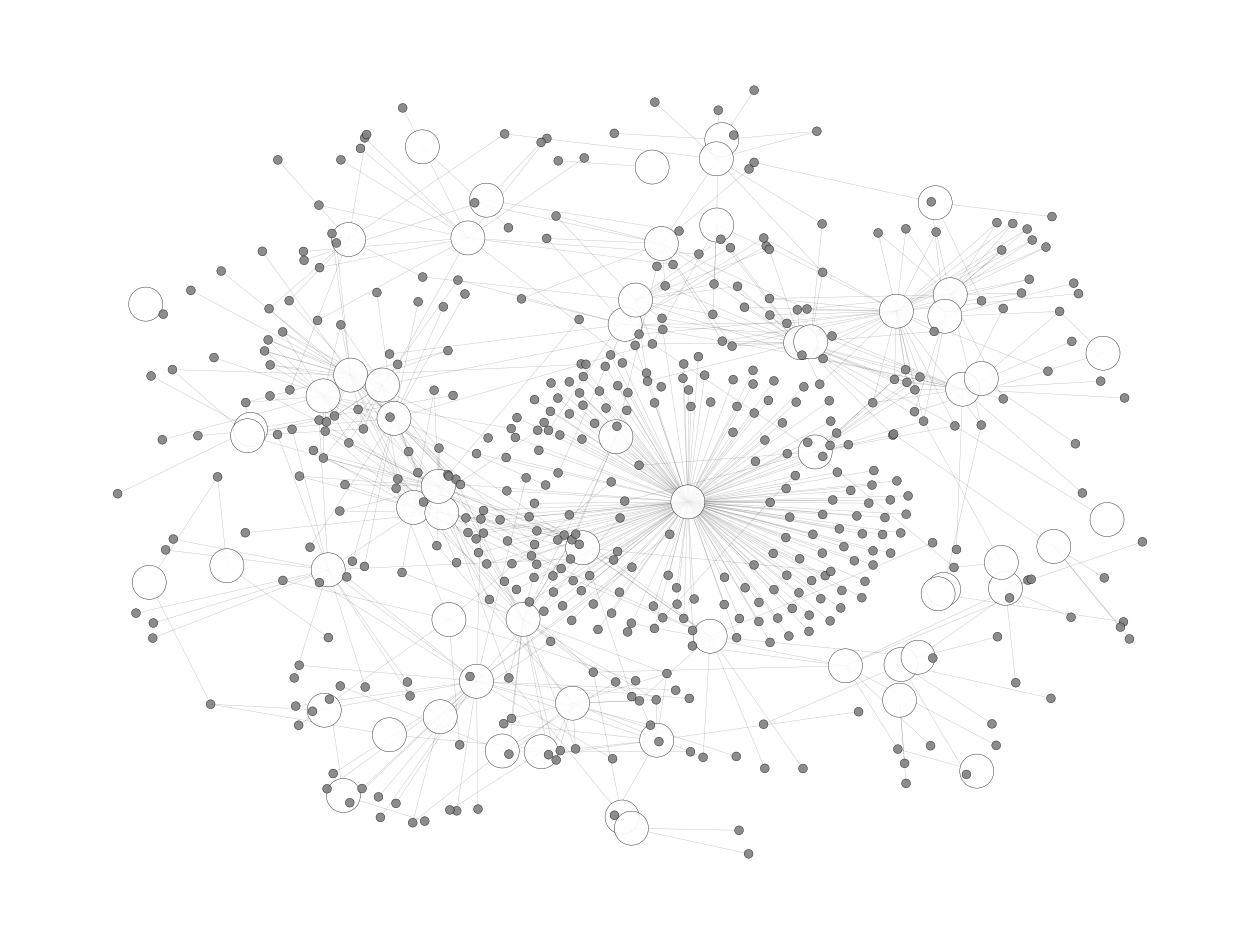}
        \caption{MPTM Gemini-2.0}
    \end{subfigure}
    \begin{subfigure}[t]{0.24\textwidth}
        \includegraphics[width=\linewidth]{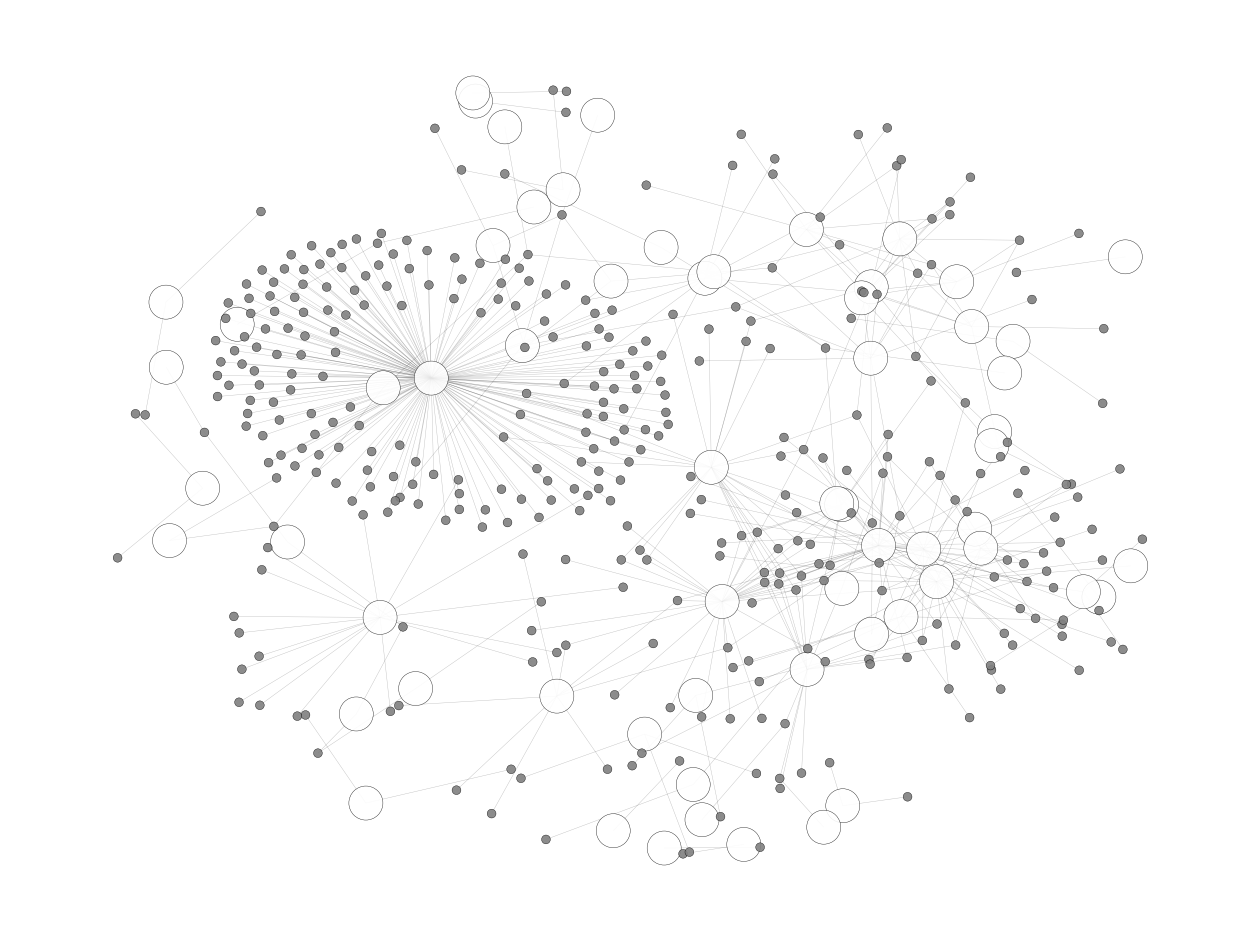}
        \caption{MPTM GPT-4o}
    \end{subfigure}

    \caption{Bipartite graphs between named entities (white nodes) and posts (grey nodes) in elections dataset [(a),(b),(c),(d)] and influencers dataset [(e),(f),(g),(h)]. All graphs are generated using MPTM prompting by three LLMs.}
    \label{fig:graphs_mp}
\end{figure}

\begin{figure}[H]
    \centering
   \begin{subfigure}[t]{0.24\textwidth}
        \includegraphics[width=\linewidth]{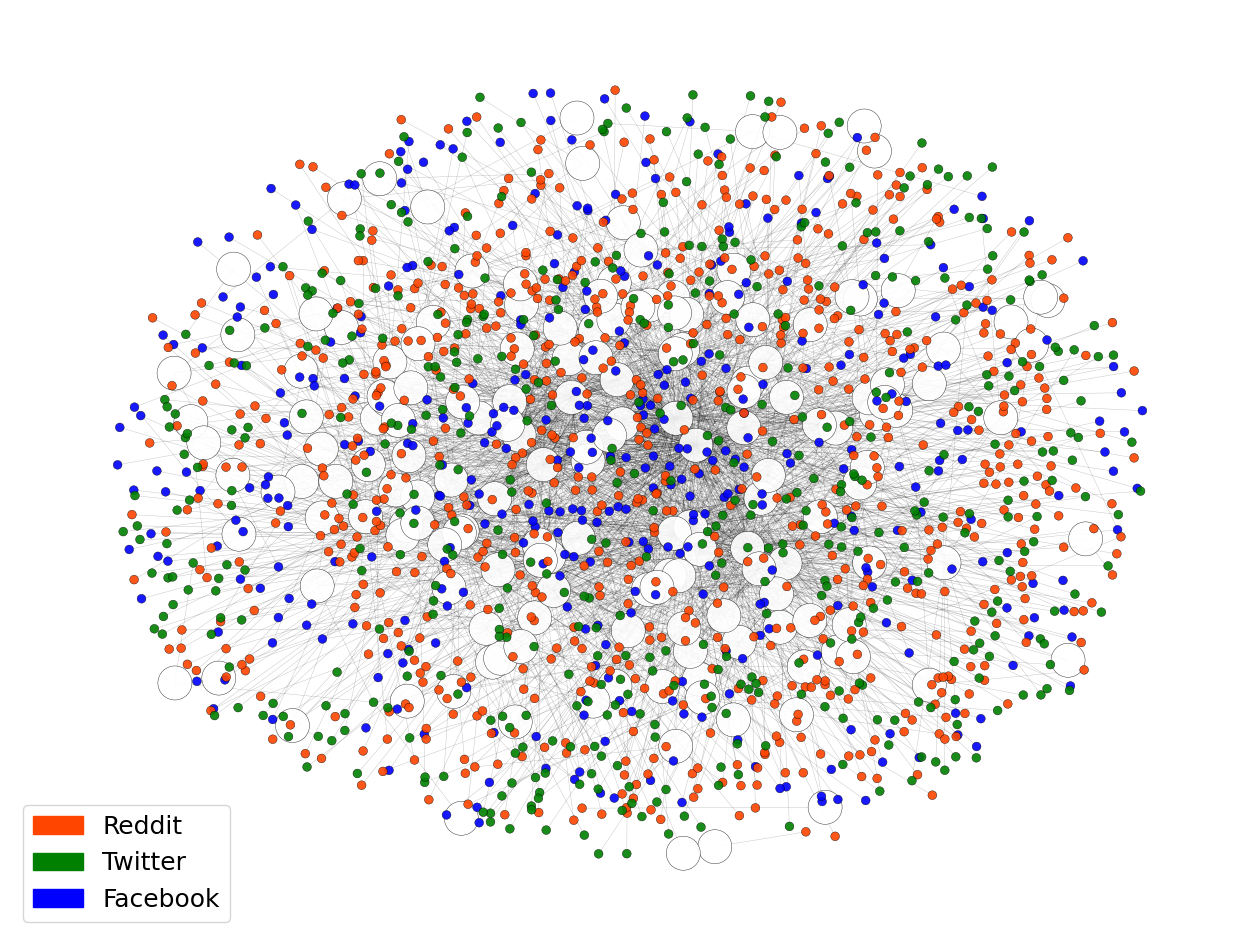}
        \caption{PP Real}
    \end{subfigure}   
    \begin{subfigure}[t]{0.24\textwidth}
        \includegraphics[width=\linewidth]{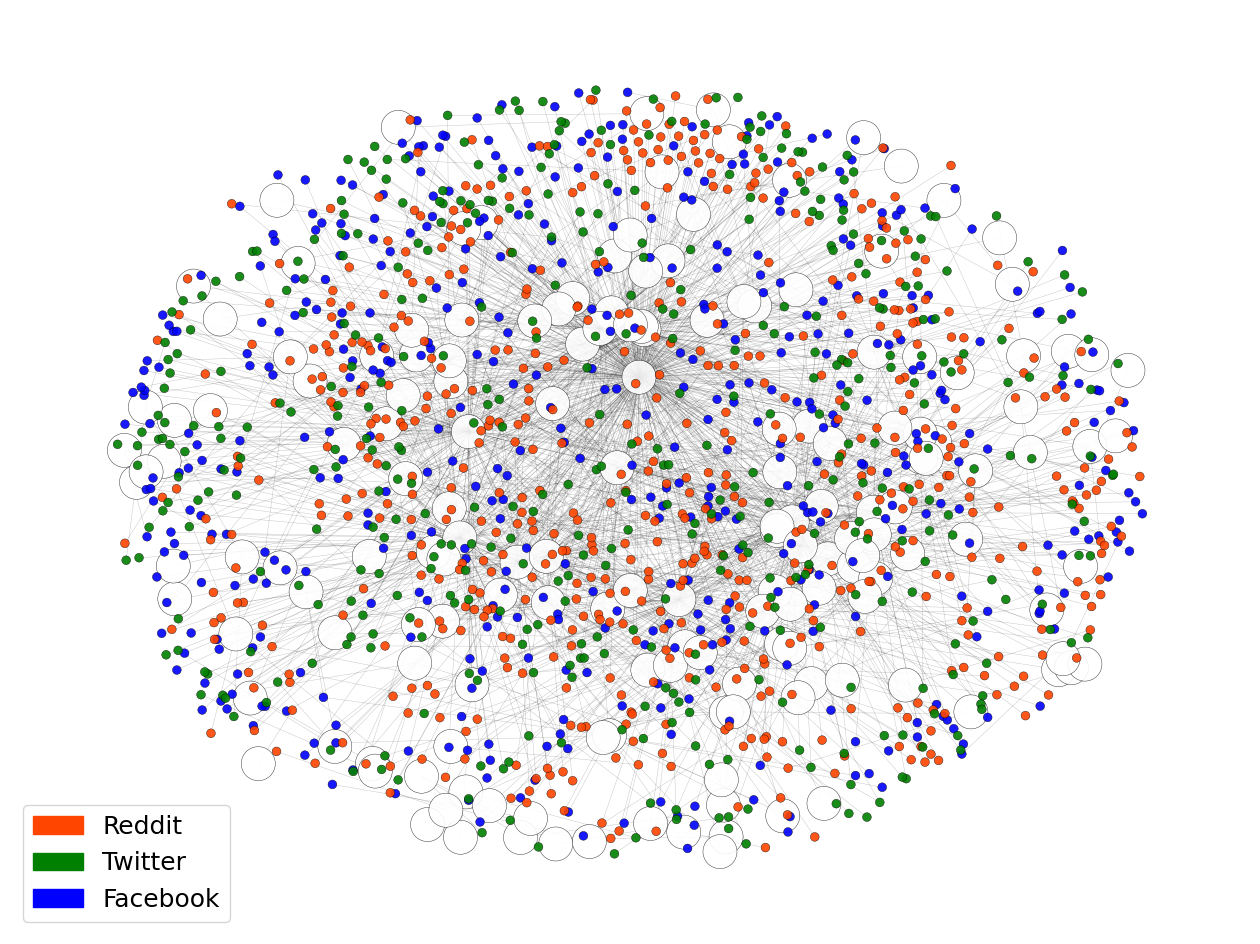}
        \caption{PP Claude-3.5}
    \end{subfigure}
    \begin{subfigure}[t]{0.24\textwidth}
        \includegraphics[width=\linewidth]{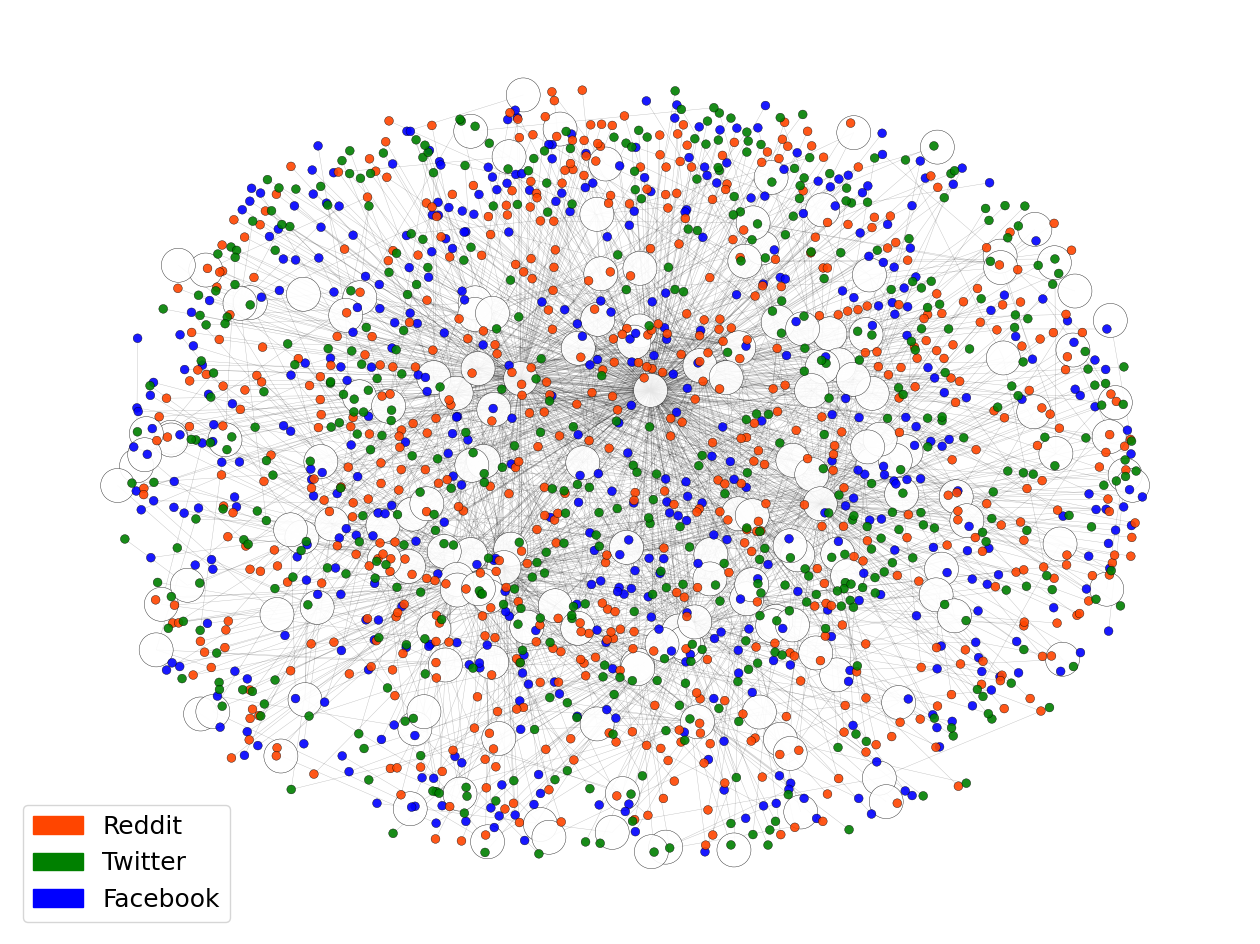}
        \caption{PP Gemini-2.0}
    \end{subfigure}
    \begin{subfigure}[t]{0.24\textwidth}
        \includegraphics[width=\linewidth]{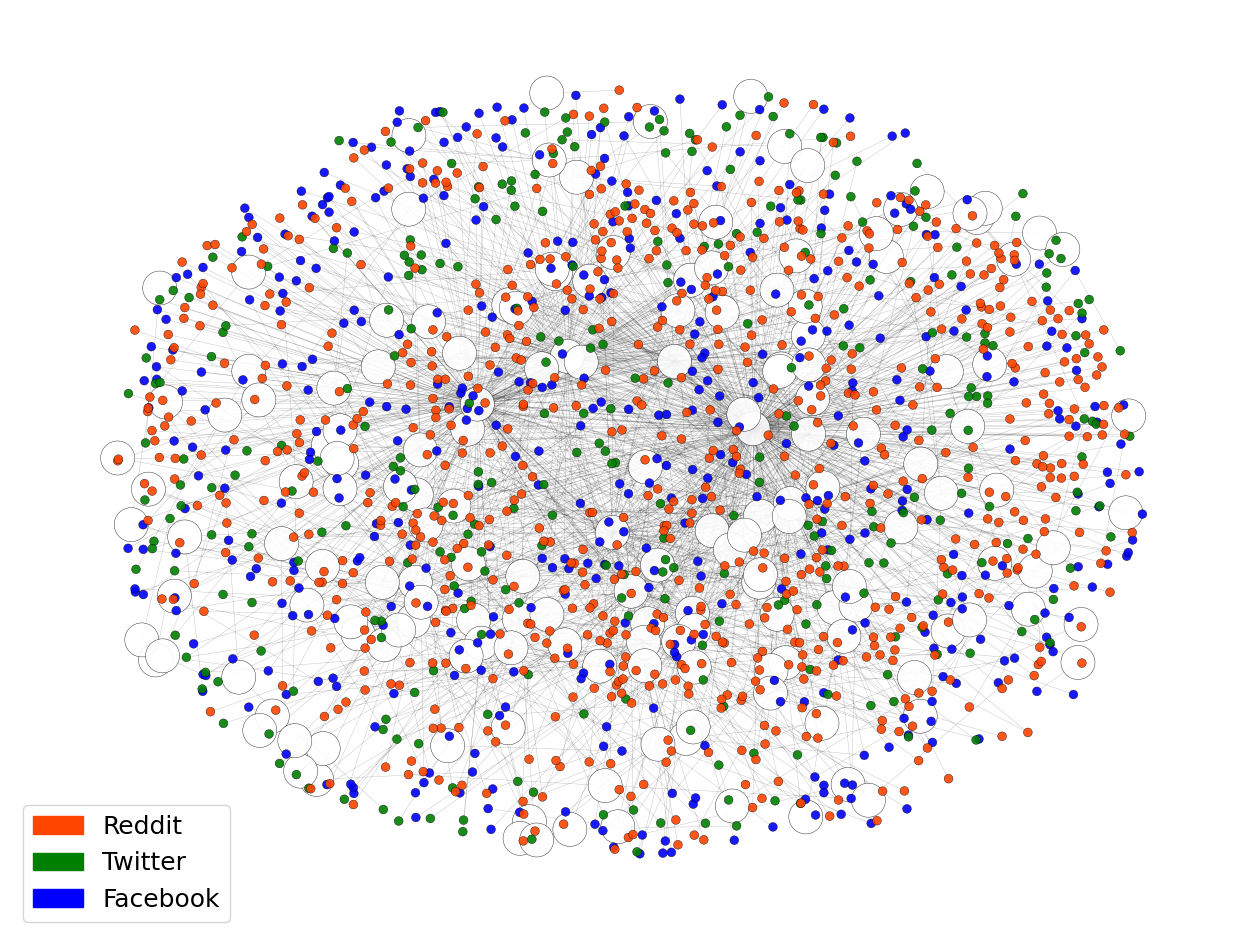}
        \caption{PP GPT-4o}
    \end{subfigure}

    \begin{subfigure}[t]{0.24\textwidth}
        \includegraphics[width=\linewidth]{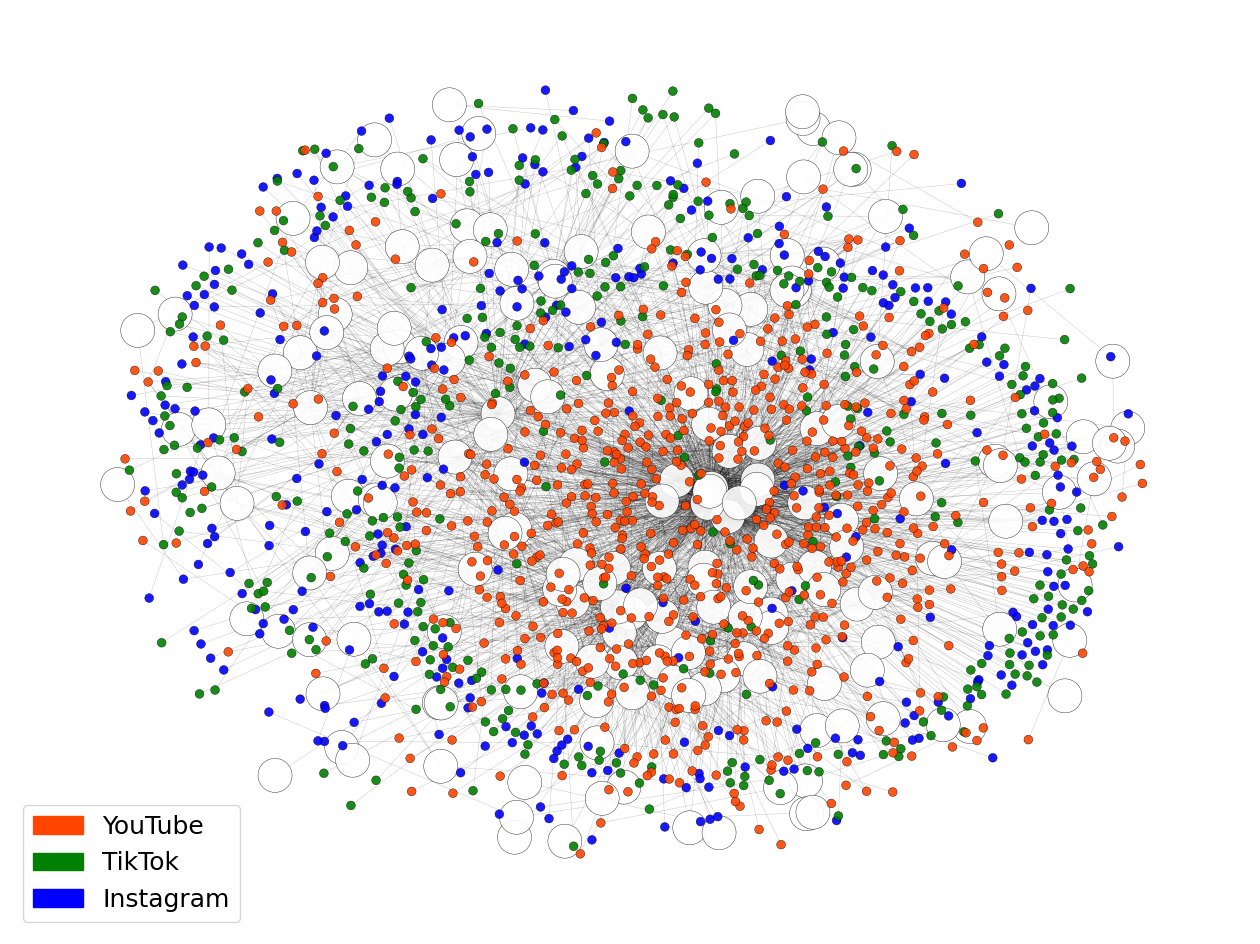}
        \caption{PP Real}
    \end{subfigure}
    \begin{subfigure}[t]{0.24\textwidth}
        \includegraphics[width=\linewidth]{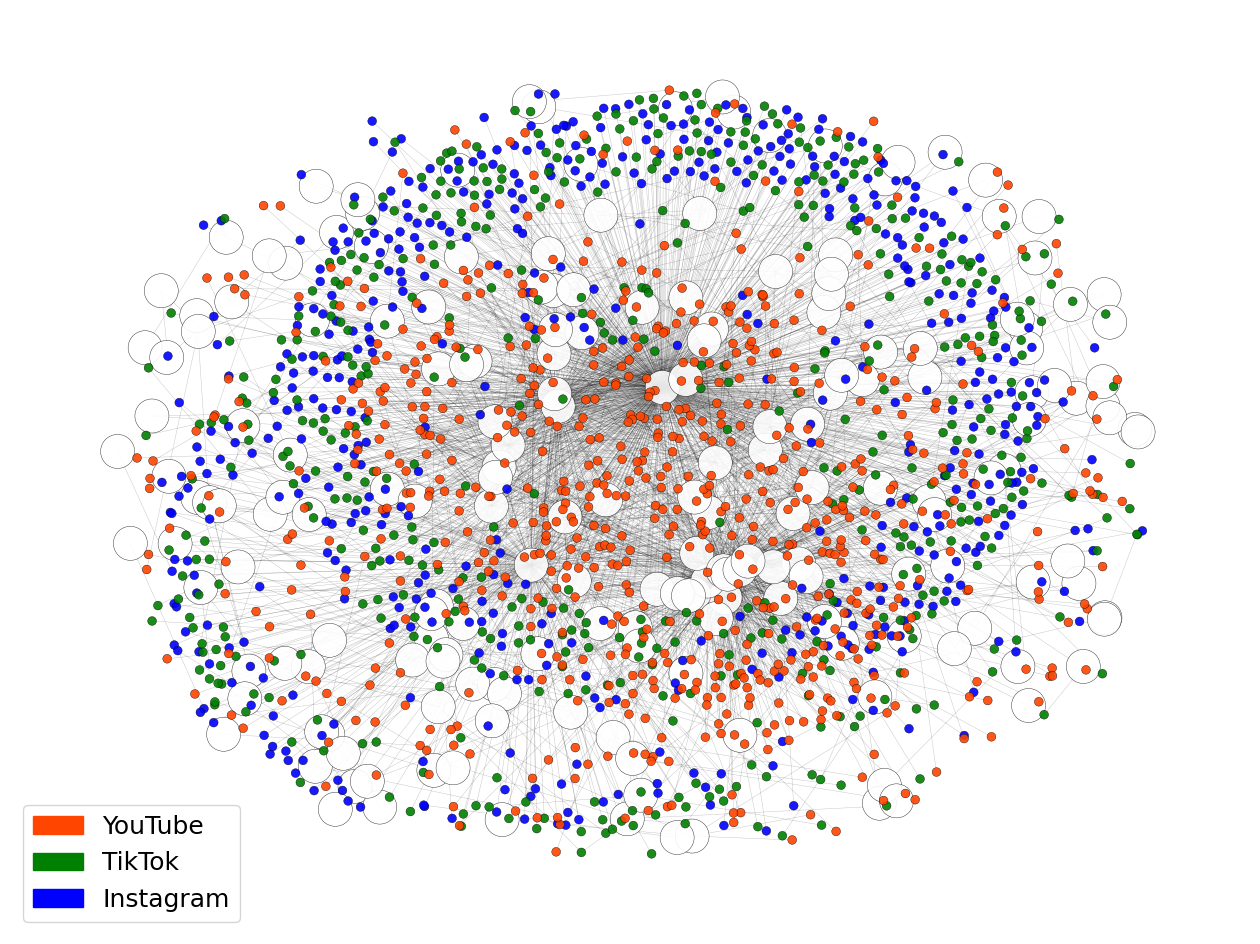}
        \caption{PP Claude-3.5}
    \end{subfigure}
    \begin{subfigure}[t]{0.24\textwidth}
        \includegraphics[width=\linewidth]{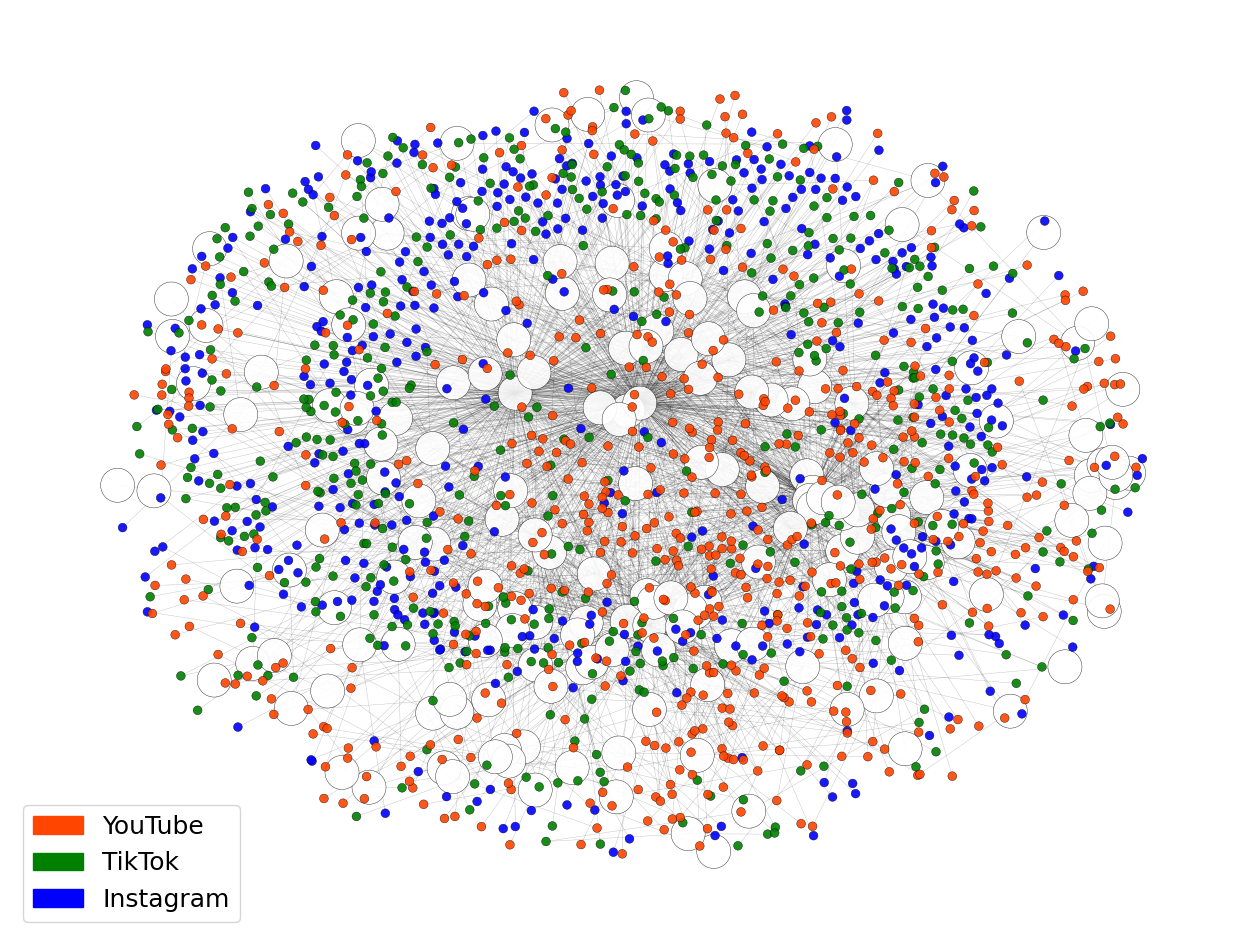}
        \caption{PP Gemini-2.0}
    \end{subfigure}
    \begin{subfigure}[t]{0.24\textwidth}
        \includegraphics[width=\linewidth]{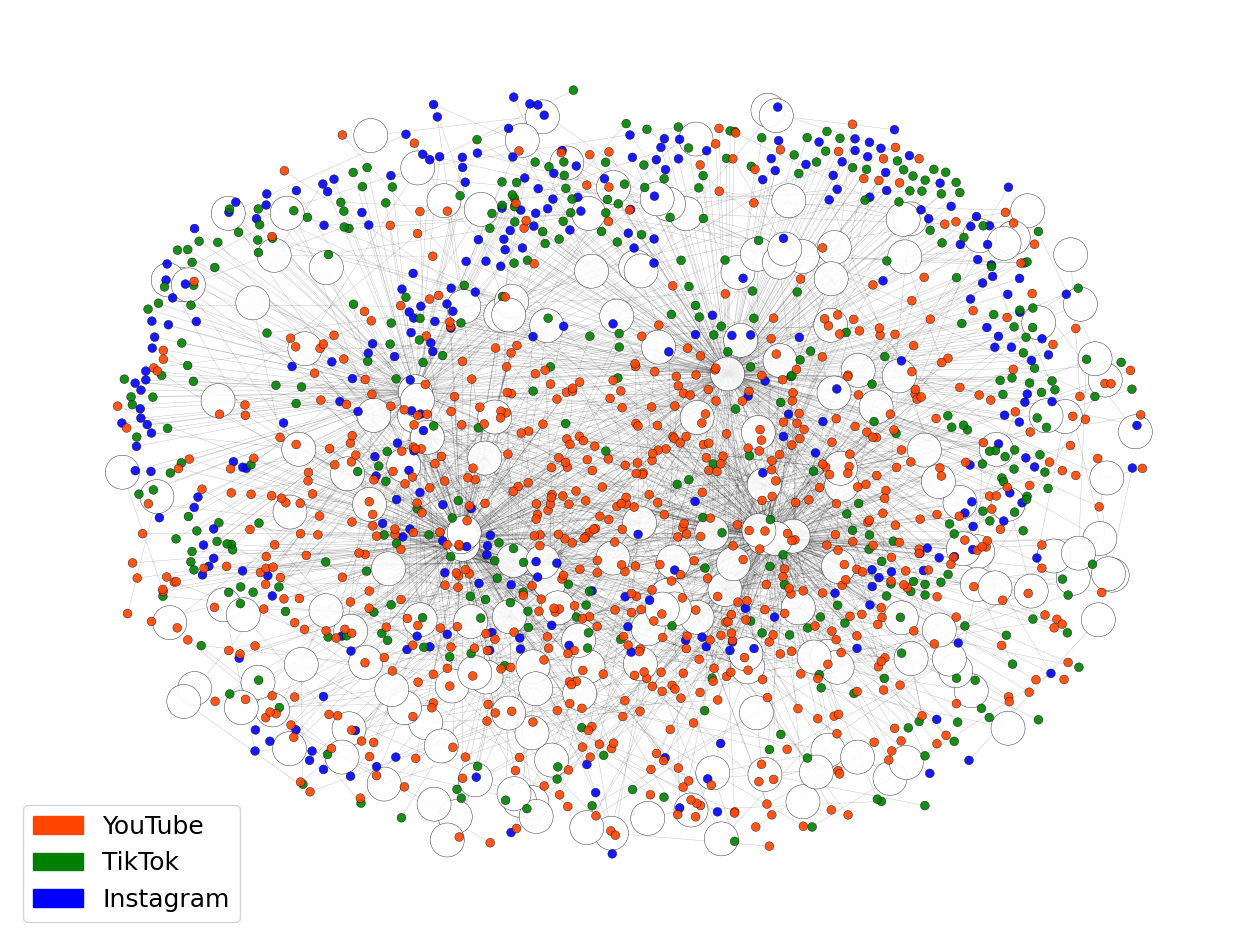}
        \caption{PP GPT-4o}
    \end{subfigure}
    \caption{Bipartite graphs between named entities (white nodes) and posts (in different color nodes depending upon platforms) mentioning those named entities in elections dataset [(a),(b),(c),(d)] and influencers dataset [(e),(f),(g),(h)]. All graphs are generated using Per-Platform prompting by three LLMs. Nodes with red, blue, and green represent posts in Reddit, Twitter, and Facebook, respectively, in elections dataset and YouTube, TikTok, and Instagram, respectively in influencers dataset.}
    \label{fig:graphs_pp}
\end{figure}

We propose a multi-platform metric that measures the distribution of named entities across different platforms. The analysis follows: (1) Named Entity Recognition (NER) is performed on both real and synthetic datasets. (2) Bipartite graphs are constructed for each dataset, depicted in Fig.~\ref{fig:graphs_mp} and Fig.~\ref{fig:graphs_pp} for multi-platform and per-platform prompting, respectively. Common entities are represented as white nodes and posts with different colors depending on the platforms where they were posted. An edge means that the post mentions the named entity. In Tables~\ref{tab:graph_stats_mp} and~\ref{tab:graph_stats_pp}, we present the node count and edge count in these bipartite graphs. (3) The degree distribution of named entity (white) nodes is computed for all graphs and compared with each other. 

In Fig.~\ref{fig:graphs_mp}, we plot the bipartite graphs of named entities in synthetic datasets generated by all LLMs in multi-platform prompting. White nodes represent the named entities, and grey nodes represent posts that mention those named entities. 
In Table~\ref{tab:graph_stats_mp}, we specify these bipartite graphs' node counts and edge counts. In both the elections dataset and the influencers dataset, we find that commonly named entities are spread across more posts in the synthetic dataset generated by the Claude-3.5 model and several posts in GPT-4o generated content. Gemini-2.0 gives mixed results, underestimating in elections and overestimating in influencers datasets. 
Claude-3.5 also produces both datasets with more mentions (more edges), whereas GPT-4o generates data with fewer mentions. Gemini-2.0 gives fewer mentions in elections and more mentions in the influencers dataset.

\begin{table}[H]
\centering
\caption{Named Entity Bipartite Graph Statistics - MPTM prompting. Named Entities are white nodes and posts are grey nodes.}
\footnotesize
\begin{tabular}{@{}lcccccc@{}}
\toprule
& \multicolumn{3}{c}{\textbf{US Election Dataset}} & \multicolumn{3}{c}{\textbf{Influencers Dataset}}\\
\hline
 & \multicolumn{2}{c}{\textbf{Node Count}} & \multicolumn{1}{c}{\textbf{Edge Count}} & \multicolumn{2}{c}{\textbf{Node Count}} & \multicolumn{1}{c}{\textbf{Edge Count}} \\
 \textbf{LLM}  & \textbf{white} & \textbf{grey} & \textbf{white-grey} & \textbf{white} & \textbf{grey} & \textbf{white-grey}\\
\midrule
 Sample Pool & 84 & 533 & 824 & 62 & 390 & 615 \\
 Claude-3.5 & 84 & 588 & 805 & 62 & 409 & 621 \\
 Gemini-2.0 & 84 & 491 & 623 & 62 & 442 & 652 \\
 GPT-4o     & 84 & 519 & 668 & 62 & 360 & 481 \\
\bottomrule
\end{tabular}
\label{tab:graph_stats_mp}
\end{table}

\begin{table}[H]
\centering
\caption{Named Entity Bipartite Graph Statistics - Per-Platform prompting. Named Entities are white nodes and posts are either red, blue or green nodes depending upon platforms. Elections: Reddit(Rd)=red, Twitter(Tw)=green, Facebook(Fb)=blue and Influencers: YouTube(Yt)=red, TikTok(Tk)=green, Instagram(In): blue.}
\footnotesize
\begin{tabular}{@{}lccccccc@{}}
\toprule
& \multicolumn{7}{c}{\textbf{US Election Dataset}} \\
\hline
  & \multicolumn{4}{c}{\textbf{Node Count}} & \multicolumn{3}{c}{\textbf{Edge Count}} \\
  \textbf{LLM}  & \textbf{white} & \textbf{red} & \textbf{green} & \textbf{blue} & \textbf{W-R} & \textbf{W-G} & \textbf{W-B} \\
  & & \textbf{Rd} & \textbf{Tw} & \textbf{Fb} & & & \\
\midrule
Sample Pool & 169 & 784 & 495 & 403 & 1243 & 737 & 798 \\
Claude-3.5 & 169 & 607 & 440 & 487 & 772 & 510 & 560 \\
Gemini-2.0 & 169 & 645 & 515 & 505 & 794 & 608 & 632\\
GPT-4o     & 169 & 790 & 315 & 472 & 917 & 372 & 580 \\
\midrule
& \multicolumn{7}{c}{\textbf{Dutch Influencers Dataset}} \\
\hline
  & \multicolumn{4}{c}{\textbf{Node Count}} & \multicolumn{3}{c}{\textbf{Edge Count}} \\
 \textbf{LLM}  & \textbf{white} & \textbf{red} & \textbf{green} & \textbf{blue} & \textbf{W-R} & \textbf{W-G} & \textbf{W-B} \\
   & & \textbf{Yt} & \textbf{Tk} & \textbf{In} & & & \\
\midrule
Sample Pool & 197 & 754 & 417 & 374 & 2507 & 535 & 430 \\
Claude-3.5 & 197 & 731 & 516 & 535 & 1684 & 549 & 550 \\
Gemini-2.0 & 197 & 652 & 499 & 501 & 1426 & 610 & 536\\
GPT-4o     & 197 & 740 & 377 & 308 & 1546 & 410 & 325\\
\bottomrule
\end{tabular}
\label{tab:graph_stats_pp}
\end{table}
In Fig.~\ref{fig:graphs_pp}, we plot the bipartite graphs of named entities in synthetic datasets generated by all LLMs in per-platform prompting. White nodes represent the named entities. Nodes with red, green, and blue represent posts in Reddit, Twitter, and Facebook, respectively, in the elections dataset, and YouTube, TikTok, and Instagram, respectively, in the influencers dataset. 
In Table~\ref{tab:graph_stats_pp}, we specify these bipartite graphs' node counts and edge counts. In both the elections dataset and influencers dataset, we find posts on Reddit and YouTube mentioning common entities that almost match those in real counterparts (Sample Pool). This could be attributed to the fact that posts are longest on Reddit and YouTube, which allows LLMs to generate posts with high fidelity.
GPT-4o underrepresents posts on both Twitter and TikTok. Facebook posts are overestimated in synthetic datasets generated by LLMs. Looking at the edge distributions, we find that all LLMs fail to generate mentions (edges) to common named entities in their posts in proportion to those in real datasets.  

\begin{figure}[htbp]
    \centering
    \begin{subfigure}[b]{0.48\textwidth}
        \includegraphics[width=\textwidth]{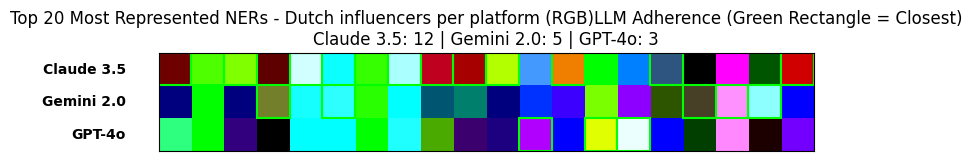}
        \caption{Top 20 - Dutch Influencers (Per Platform)}
    \end{subfigure}
    \hfill
    \begin{subfigure}[b]{0.48\textwidth}
        \includegraphics[width=\textwidth]{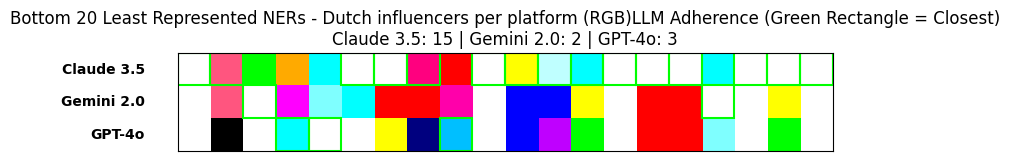}
        \caption{Bottom 20 - Dutch Influencers (Per Platform)}
    \end{subfigure}

    \begin{subfigure}[b]{0.48\textwidth}
        \includegraphics[width=\textwidth]{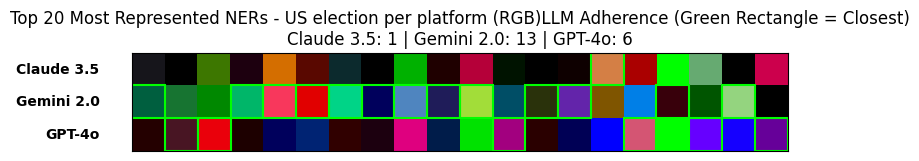}
        \caption{Top 20 - US Election (Per Platform)}
    \end{subfigure}
    \hfill
    \begin{subfigure}[b]{0.48\textwidth}
        \includegraphics[width=\textwidth]{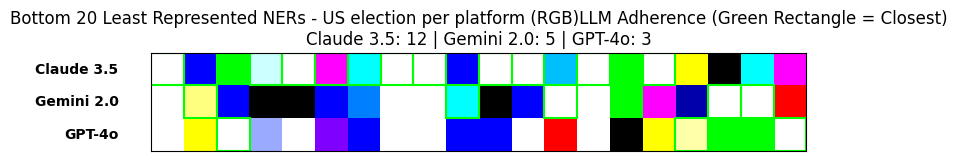}
        \caption{Bottom 20 - US Election (Per Platform)}
    \end{subfigure}
    \begin{subfigure}[b]{0.48\textwidth}
        \includegraphics[width=\textwidth]{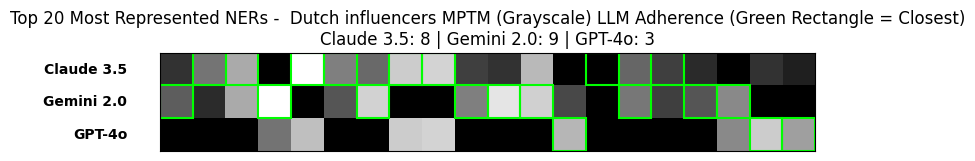}
        \caption{Top 20 - Dutch Influencers (MPTM)}
    \end{subfigure}
    \hfill
    \begin{subfigure}[b]{0.48\textwidth}
        \includegraphics[width=\textwidth]{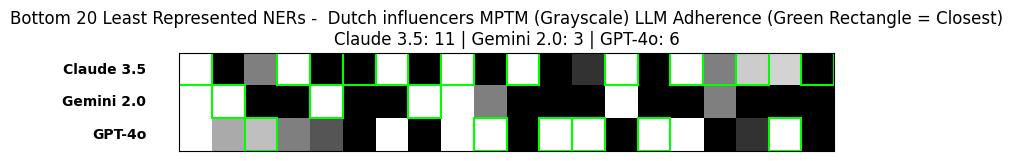}
        \caption{Bottom 20 - Dutch Influencers (MPTM)}
    \end{subfigure}

    \begin{subfigure}[b]{0.48\textwidth}
        \includegraphics[width=\textwidth]{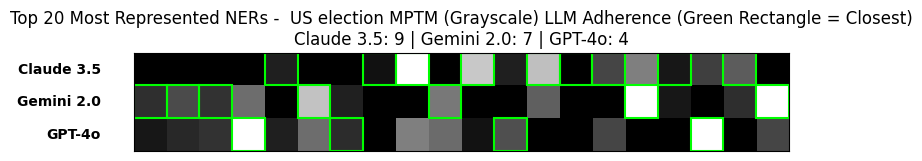}
        \caption{Top 20 - US Election (MPTM)}
    \end{subfigure}
    \hfill
    \begin{subfigure}[b]{0.48\textwidth}
        \includegraphics[width=\textwidth]{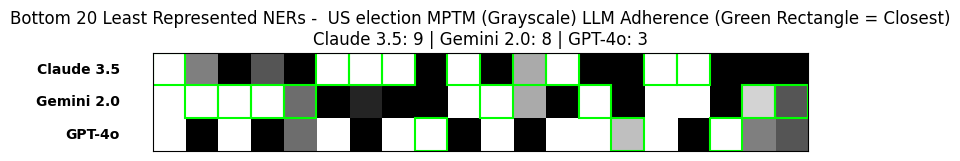}
        \caption{Bottom 20 - US Election (MPTM)}
    \end{subfigure}

    \caption{LLM adherence to NER distributions across datasets. Each subplot shows how closely Claude 3.5, Gemini 2.0, and GPT-4o match the ground truth for the 20 most and least represented entities. Green rectangles indicate the closest match.}
    \label{fig:ner-adherence}
\end{figure}

In Fig.~\ref{fig:ner-adherence}, we depict the adherence of LLMs to named entity distributions across different platforms in the two datasets. 
We show top-20 (most frequently occurring) and bottom-20 (least frequently occurring) common named entities. 
Each named entity is represented as a white node, say $v_w$. In per-platform settings, the degree ($deg(v_w)$) of each white node is represented by a degree vector ($[v_w^{p1},v_w^{p2},v_w^{p3}]$) of three dimensions which can be understood as connections to three different platforms. For each of top-20 and bottom-20 common entities (white nodes) in real and synthetic, we compute absolute difference between their degree vectors i.e. $|[v_w^{p1},v_w^{p2},v_w^{p3}]_{real} - [v_w^{p1},v_w^{p2},v_w^{p3}]_{synthetic}|$. The smaller the difference, the more adherence there is to synthetic content with real content. 
To present this absolute difference in a color palette, we rescale vector difference in the red, green, and blue color ranges between $[0,0,0]$ (black) to $[255,255,255]$ (white). 
We flip the color values so that named entities in synthetic data that adheres to real data is represented with brighter color rather than darker color. 
Fig.~\ref{fig:ner-adherence} shows that we have brighter colors in the bottom-20 common named entities and darker colors in the top-20 common named entities in both per-platform and multi-platform prompt outputs. 
This means that named entities in synthetic data generated by all LLMs adhere to real data in less frequent common named entities and struggle when the common named entities are more frequent.

%% file: sections/summary.tex
\section{Summary and Discussions}
This paper investigates using three LLMs (Claude-3.5, Gemini-2.0, and GPT-4o) to generate social media datasets in multi-platform scenarios. We propose a new prompt technique, MPTM prompting, which gives topically relevant examples to LLMs for few-shot prompting. We evaluate synthetic post generation in two diverse datasets. The first dataset includes posts from Facebook, Reddit, and Twitter related to the US midterm 2022 elections, while the second comprises posts from TikTok, Instagram, and YouTube by a group of registered Dutch influencers.
This work is an extension to our previous work~\cite{tari2024leveraging} where we employed two prompting strategies: one where chatGPT was prompted to generate posts similar to given examples without specifying the platform, and another where it was prompted to generate platform-specific posts. 
In this work, we compare MPTM (multi-platform) prompting results with our per-platform prompting approach from our previous work. 
We assessed the quality of the generated datasets based on lexical features (hashtags, URLs, emojis, and user tags), sentiment, topics, embedding similarity, and named entities.
Our findings indicate that the synthetic datasets generally display high fidelity across various metrics. The synthetic posts notably preserve key social media lexical features such as emojis and hashtags and maintain semantic similarities with the original datasets, demonstrated through topic analysis and embedding similarity. However, neither prompting strategy consistently outperformed the other.

We also compare the relative advantages of top of the art commercial LLMs for this task. 
GPT-4o underestimates hashtag usage compared to GPT-3.5-Turbo, while Claude-3.5 and Gemini-2.0 generally overproduce hashtags. For mentions, Claude-3.5 overestimates and Gemini-2.0 underestimates, whereas GPT-4o aligns more closely with real data in multi-platform settings. None of the LLMs effectively replicate URL usage, especially the high volume seen in YouTube descriptions. GPT-4o and Claude-3.5 generate more emojis than GPT-3.5-Turbo, while Gemini-2.0 underproduces them.
Reddit exhibits the highest negativity in real data. GPT-4o underestimates negative sentiment, while Claude-3.5 and Gemini-2.0 tend to overestimate it. All LLMs produce more positive and fewer neutral posts across platforms, regardless of the prompting style.
Topic overlap is stronger in the synthetic influencers dataset than in the election dataset, with Facebook topics better represented across all LLMs. Gemini-2.0 performs best for topic overlap on TikTok and YouTube. While election-related topics are well captured, influencer-related datasets struggle due to their high topical diversity.
Embedding clusters from MPTM-prompted synthetic data show better alignment with real data in the elections dataset than in the influencers dataset. LLMs better replicate infrequent named entities but struggle with frequently occurring ones.

The study has limitations that can be explored in future research. First, our results are based solely on the output of popular LLMs as first quarter of 2025, setting a benchmark for future studies that might replicate our approach with different LLMs. 
Second, in this work we do not address privacy and legal concerns related to releasing synthetic datasets.
Third, while focusing on a downstream task is beyond the scope of this paper, we acknowledge that measuring task-specific utility of synthetic data is a necessary complementary metric on the path towards reliable adoption of synthetic data use in computational social sciences. 

This research demonstrates that current technology makes generating synthetic social media datasets spanning multiple platforms feasible. Improved prompting techniques and post-processing could enhance dataset fidelity. Realistic multi-platform synthetic datasets can promote reproducibility in social media research and offer resilience against restrictive platform policies, facilitating access to data for researchers globally.